\documentclass[10pt,twocolumn,letterpaper]{article}

\usepackage[pagenumbers]{cvpr} %

\usepackage{booktabs} %
\usepackage{multirow}
\usepackage{makecell}
\usepackage[title]{appendix}
\usepackage{orcidlink}
\usepackage{booktabs}
\usepackage{multirow}
\usepackage{makecell}
\usepackage{amsmath}
\usepackage{amssymb}

\definecolor{MyDarkBlue}{rgb}{0,0.08,1}
\definecolor{MyDarkGreen}{rgb}{0.02,0.6,0.02}
\definecolor{MyDarkRed}{rgb}{0.8,0.02,0.02}
\definecolor{MyDarkOrange}{rgb}{0.40,0.2,0.02}
\definecolor{MyPurple}{RGB}{111,0,255}
\definecolor{MyRed}{rgb}{1.0,0.0,0.0}
\definecolor{MyGold}{rgb}{0.75,0.6,0.12}
\definecolor{MyDarkgray}{rgb}{0.66, 0.66, 0.66}
\definecolor{MyDarkCyan}{rgb}{0.05, 0.55, 0.45}
\definecolor{MyBlack}{rgb}{0., 0., 0.}
\definecolor{MyMagenta}{rgb}{1., 0., 1.}
\definecolor{BerkeleyYellow}{RGB}{255,204,41}
\definecolor{BerkeleyLightBlue}{RGB}{94,146,221}
\definecolor{BkDarkBlue}{rgb}{.05,.07,.353}

\newcommand{\maxwellchange}[1]{\textcolor{MyBlack}{#1}}

\newcommand{\secondmaxwellchange}[1]{\textcolor{MyBlack}{#1}}

\newcommand{\myparagraph}[1]{\vspace{2pt} \noindent \textbf{#1} \ }

\newcommand*\widebar[1]{\@ifnextchar^{{\wide@bar{#1}{0}}}{\wide@bar{#1}{1}}}

\def\x{{\mathbf{x}}}
\def\c{{\mathbf{c}}}

\newcommand{\reffig}[1]{Figure~\ref{fig:#1}}

\newcommand{\refapp}[1]{Appendix~\ref{sec:#1}}

\newcommand{\ignorethis}[1]{}

\newcommand*{\menlo}{\fontfamily{lmtt}\fontsize{9}{9}\selectfont }

\newcommand{\methodname}{\texttt{Pair\;Customization}}

\newcommand{\wordstyle}
{<desc> }

\begin{document}

\title{Customizing Text-to-Image Models with a Single Image Pair}

\author{Maxwell Jones\textsuperscript{1}
\qquad
Sheng-Yu Wang\textsuperscript{1}
\qquad
Nupur Kumari\textsuperscript{1}
\qquad
David Bau\textsuperscript{2}
\qquad 
Jun-Yan Zhu\textsuperscript{1}\\
\textsuperscript{1}Carnegie Mellon University
\qquad
\textsuperscript{2}Northeastern University
}
\twocolumn[{%
\renewcommand\twocolumn[1][]{#1}%

\maketitle

\begin{center}
    \centering
    \includegraphics[width=\linewidth]{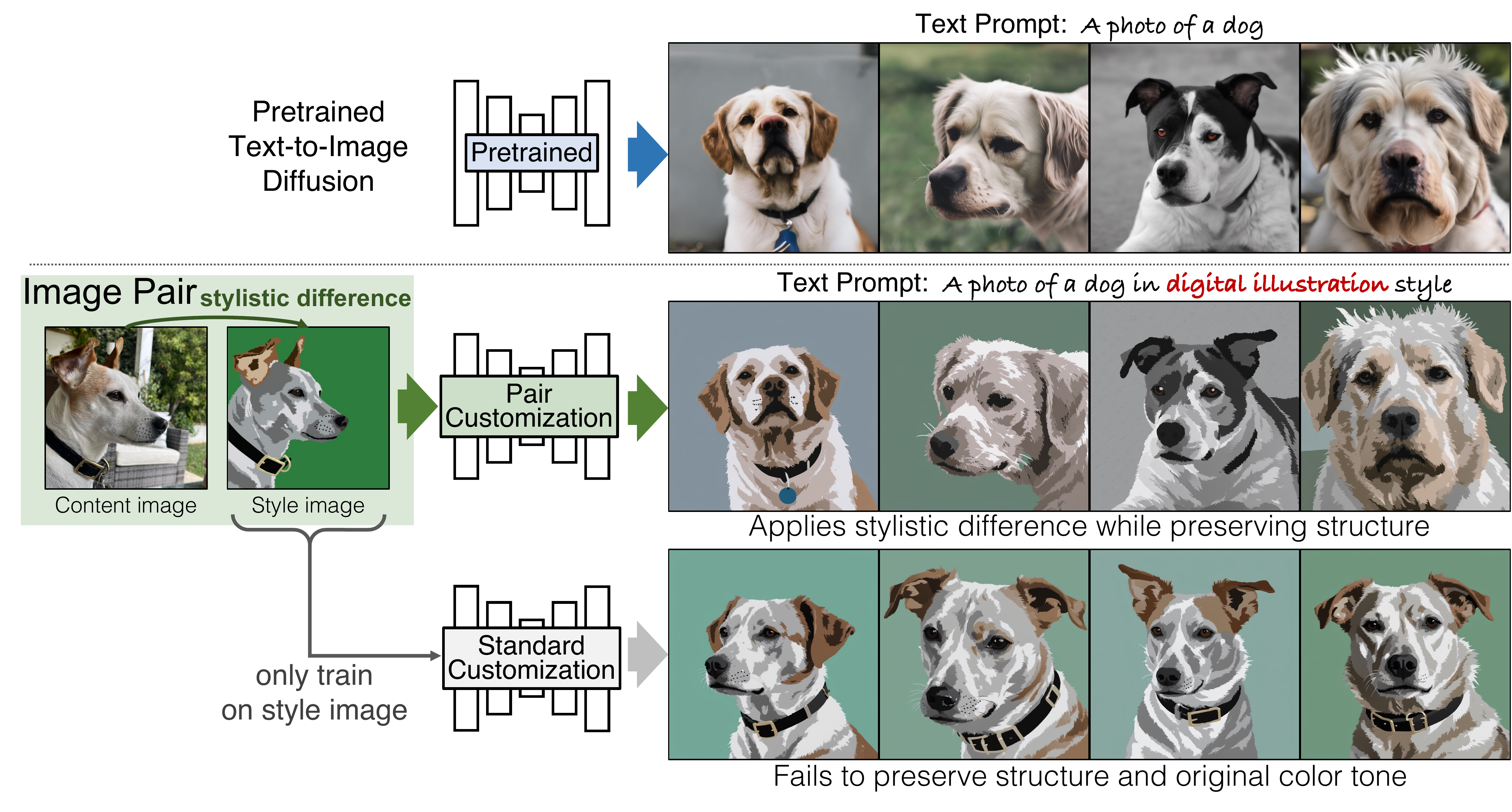}
    \vspace{-7pt}
    \captionof{figure}{
    Given a \emph{single} image pair, we present \methodname, a method for customizing a pre-trained text-to-image model and learning a new style from the image pair's stylistic difference. Our method can apply the learned stylistic difference to new input images while preserving the input structure. Compared to Dreambooth LoRA\cite{hu2022lora, loraimplementation}, a standard customization method that solely uses style images, our method effectively disentangles style and content, resulting in better structure, color preservation, and style application. Style image credit: \href{https://www.instagram.com/parkhouse_art/}{Jack Parkhouse}.
    }
    \label{fig:teaser}
\end{center}
}]
 \maketitle
 
\begin{abstract}

Art reinterpretation is the practice of creating a variation of a reference work, making a paired artwork that exhibits a distinct artistic style. 
We ask if such an image pair can be used to customize a generative model to capture the demonstrated stylistic difference.
We propose \methodname, a new customization method that learns stylistic difference from a \emph{single} image pair and then applies the acquired style to the generation process. %
Unlike existing methods that learn to mimic a single concept from a collection of images, our method captures the stylistic difference between paired images. This allows us to apply a stylistic change without overfitting to the specific image content in the examples.  To address this new task, we employ a joint optimization method that explicitly separates the style and content into distinct LoRA weight spaces. We optimize these style and content weights to reproduce the style and content images while encouraging their orthogonality. During inference, we modify the diffusion process via a new style guidance based on our learned weights. 
Both qualitative and quantitative experiments show that our method can effectively learn style while avoiding overfitting to image content, highlighting the potential of modeling such stylistic differences from a single image pair.

\end{abstract}

\section{Introduction}
\label{sec:intro}

Artistic works are often inspired by a reference image, a recurring scene, or even a previous piece of art~\cite{artreinterpretation}.
Such creations involve re-interpreting an original composition in the artist's unique style. A notable example is Van Gogh's \emph{Repetitions}~\cite{artrepetition}, in which the artist created multiple versions of the same scenes with his distinctive expressiveness, including adaptations of other artists' work. Such sets of variations allow close comparison of stylized art to a reference image, providing unique insights into an artist's detailed techniques and choices.

In our work, we explore how such \textit{content-style image pairs} can be used to customize a generative model to capture the demonstrated stylistic difference. Our goal is to customize a pre-trained generative model to synthesize stylized images, distilling the essence of the style from as few as a single pair without fixating on specific content. We wish to create a model capable of re-interpreting a variety of different content in the style demonstrated by the paired variation.

Prior works on model customization/personalization~\cite{gal2022image,ruiz2023dreambooth,kumari2023multi} take one or a few images of a single concept to customize large-scale text-to-image models~\cite{ramesh2022hierarchical,rombach2022high}. While they aim to learn styles without using pairs, the generated samples from these customized models often resemble the training images' content, such as specific objects, persons, and scene layouts. %
In Figure~\ref{fig:teaser}, we observe that standard single-image customization ($3^{\text{rd}}$ row) \maxwellchange{alters the subject, color tone, and pose of} the original image ($1^{\text{st}}$ row). These issues arise because the artistic intent is difficult to discern from a single image: unlike image pairs that can demonstrate a style through contrasts, a singleton example will always intertwine choices of both style and content. Due to this ambiguity, the model fails to capture the artistic style accurately and, in some cases, overfits and generates the subject-specific details rather than the style, as shown in Figure~\ref{fig:main_result}.

On the other hand, our {\methodname} 
method exploits the contrast between image pairs to generate pairwise consistent images while better disentangling style and content.  In Figure~\ref{fig:teaser} ($2^{\text{nd}}$ row), our method accurately follows the given style, \maxwellchange{turning the background into a single color matching the original background and preserving the identity and pose for each dog.}
Our method achieves this by disentangling the intended style from the image pair.

Our new customization task is challenging since text-to-image models were not initially designed to generate \emph{pairwise} content. Even when given specific text prompts like ``{\menlo a portrait}'' and ``{\menlo a portrait with Picasso style}'', a text-to-image diffusion model often struggles to generate images with consistent structure from the same noise seed. Therefore, it remains unclear how a customized model can generate stylized images while maintaining the original structure.

To address the challenges, we first propose a joint optimization method with separate sets of Low-Rank Adaptation~\cite{hu2022lora} (LoRA) weights for style and content. The optimization encourages the content LoRA to reconstruct the content image and the style LoRA to apply the style to the content. We find that the resulting style LoRA can apply the same style to other unseen content. Furthermore, we enforce row-space orthogonality~\cite{po2023orthogonal} between style and content LoRA parameters to improve style and content disentanglement. %
Next, we extend the standard classifier-free guidance method~\cite{ho2022classifier} and propose style guidance.
Style guidance integrates style LoRA predictions into the original denoising path, which aids in better content preservation and facilitates smoother control over the stylization strength. This method is more effective than the previous technique, where a customized model's strength is controlled by the magnitude of LoRA weights~\cite{dreamboothlora}. %

Our method is built upon Stable Diffusion XL~\cite{podell2023sdxl}. We experiment with various image pairs, including different categories of content (e.g., portraits, animals, landscapes) and style (e.g., paintings, digital illustrations, filters). We evaluate our method on the above single image pairs and demonstrate the advantage of our method in preserving diverse structures while applying the stylization faithfully, compared to existing customization methods. %
Our code, models, and data are available on our \href{https://paircustomization.github.io/}{webpage}.
\vspace{-5pt}

\section{Related Works}
\label{sec:formatting}

\myparagraph{Text-to-image generative models.}
Deep generative models aim to model the data distribution of a given training set~\cite{kingma2013auto,goodfellow2020generative,dinh2016density,ho2020denoising,van2016conditional,song2020score}.
Recently, large-scale text-to-image models~\cite{ramesh2022hierarchical,saharia2022photorealistic,rombach2022high,yu2022scaling,chang2023muse,kang2023scaling,sauer2023stylegan,gokaslan2023commoncanvas,podell2023sdxl,luo2023latent,balaji2022ediffi,peebles2023scalable} trained on internet-scale training data~\cite{schuhmann2021laion,kakaobrain2022coyo-700m} have shown exceptional generalization. Notably, diffusion models~\cite{ho2020denoising,song2020denoising} stand out as the most widely adopted model class. %
While existing models can generate a broad spectrum of objects and concepts, they often struggle with rare or unseen concepts. Our work focuses on teaching these models to understand and depict a new style concept. 
Conditional generative models~\cite{isola2017image,park2019semantic,zhang2023adding,mou2024t2i,li2023gligen,saharia2022palette,brooks2023instructpix2pix} learn to transform images across different domains, but the training often requires thousands to millions of image pairs. We focus on a more challenging case, where only a single image pair is available.

\myparagraph{Customizing generative models.} 
Model customization, or personalization, aims to adapt an existing generative model with additional data, with the goal of generating outputs tailored to specific user preferences. Earlier efforts mainly focus on customizing pre-trained GANs~\cite{goodfellow2020generative,karras2019style,stylegan2} for smaller datasets~\cite{stylegan2ada, diffaug, nitzan2022mystyle}, incorporating user edits~\cite{wang2021sketch,wang2022rewriting,bau2020rewriting}, or aligning with text prompts~\cite{gal2022stylegan,nitzan2023domain}. 
Recently, the focus has pivoted towards adapting large-scale text-to-image models to generate user-provided concepts, typically presented as one or a few images. Simply fine-tuning on the concept leads to overfitting. To mitigate this and enable variations via free text, several works explored different regularizations, including prior preservation~\cite{ruiz2023dreambooth,kumari2023multi}, human alignment~\cite{sohn2023styledrop}, \secondmaxwellchange{patch-based learning~\cite{zhang2023sine}}, as well as parameter update restriction, where we only update text tokens~\cite{gal2022image,daras2022multiresolution,alaluf2023neural,voynov2023p+}, attention layers~\cite{kumari2023multi,gal2023designing,han2023svdiff}, low-rank weights~\cite{hu2022lora,tewel2023keylocked,loraimplementation}, or clusters of neurons~\cite{liu2023cones}.
More recent methods focus on encoder-based approaches for faster personalization~\cite{arar2023domain,gal2023encoder,wei2023elite,li2024blip,chen2023subject,valevski2023face0,ruiz2023hyperdreambooth,ma2023subject,chen2023anydoor,shi2023instantbooth,ye2023ip}. \secondmaxwellchange{Instead of learning a single concept, several works further focus on learning multiple concepts~\cite{kumari2023multi, po2023orthogonal, gu2024mix,shah2023ziplora,avrahami2023break}. 
Other methods~\cite{ren2024customize, materzynska2023customizing} propose customizing text-to-video models to learn motion, while Guo et al.~\cite{guo2023animatediff} propose animating customized text-to-image models models by incorporating motion Low-Rank Adapter~\cite{hu2022lora} modules.}  %
Our method takes inspiration from these techniques;
however, we aim to address an inherently different task. Instead of learning concepts from an image collection, we customize the model to \textit{learn stylistic differences} from an image pair.

\myparagraph{Style and content separation.} Various past works have explored learning a style while separating it from content~\cite{tenenbaum1996separating,gatys2015neural,li2017demystifying,huang2017arbitrary,chen2016fast}. 
Our work is inspired by the seminal work Image Analogy~\cite{hertzmann2001image},  a computational paradigm that takes an image pair and applies the same translation to unseen images. Common image analogy methods include patch-wise similarity matching~\cite{hertzmann2001image,lischinski2005colorization,liao2017visual} and data-driven approaches~\cite{reed2015deepanalogy,upchurch2016z,zhu2017unpaired, park2020contrastive,visprompt,wang2023images}. Different from these, we aim to exploit the text-guided generation capabilities of large-scale models so that we can \textit{directly} use the style concept with unseen context. %
Recently, StyleDrop~\cite{sohn2023styledrop} has been proposed to learn a custom style for masked generative transformer models. Unlike StyleDrop, we do not rely on human feedback in the process. Concurrent with our work, Hertz et al.\cite{hertz2023style} introduced a method for generating images with style consistency, offering the option of using a style reference image. %
In contrast, we exploit an image pair to better discern the stylistic difference.

\section{Method}
\label{sec:method}

Our method seeks to learn a new style from a single image pair. This task is challenging, as models tend to overfit when trained on a single image, especially when generating images in the same category as the training image (e.g., a model trained and tested on dog photos). %
To reduce this overfitting, we introduce a new algorithm aimed at disentangling the structure of the subject from the style of the artwork. Specifically, we leverage the image pair to learn separate model weights for style and content.  
At inference time, we modify the standard classifier-free guidance formulation to help preserve the original image structure when applying the learned style. In this section, we give a brief overview of diffusion models, outline our design choices, and explain the final method in detail.

\subsection{Preliminary: Model Customization}
\label{sec:prelim}

\myparagraph{Diffusion models.} 
 Diffusion models\cite{ho2020denoising, song2020score, sohl2015deep},  map Gaussian noise to the image distribution through iterative denoising. Denoising is learned by reversing the forward diffusion process $\x_0,\dots,\x_T$, where image $\x_0$ is slowly \textit{diffused} to random noise $\x_T$ over $T$ timesteps, defined by $\x_t = \sqrt{\bar\alpha_t}\x_0 + \sqrt{1 - \bar\alpha_t}\epsilon$ for timestep $t \in [0, T]$. Noise $\epsilon \sim \mathcal{N}(0, I)$ is randomly sampled, and $\bar\alpha_t$ controls the noise strength. The training objective of diffusion models is to denoise any intermediate noisy image $\x_t$ via noise prediction:

 \begin{equation}
    \begin{aligned}
    \mathbb{E}_{\epsilon,\x,\c,t}\left[w_t \| \epsilon - \epsilon_{\theta}(\x_t, \c, t)\|^2\right],
\end{aligned}
\end{equation}
where $w_t$ is a time-dependent weight, $\epsilon_\theta(\cdot)$ is the denoiser that learns to predict noise, and $\mathbf{c}$ denotes extra conditioning input, such as text. At inference, the denoiser $\epsilon_\theta$ will gradually denoise random Gaussian noise into images. The resulting distribution of generated images approximates the training data distribution~\cite{ho2020denoising}.

In our work, we use Stable Diffusion XL~\cite{podell2023sdxl}, a large-scale text-to-image diffusion model built on Latent Diffusion Models~\cite{rombach2022high}. The model consists of a U-Net~\cite{ronneberger2015u} trained on the latent space of an auto-encoder, with text conditioning from two text encoders, CLIP~\cite{radford2021learning} and OpenCLIP~\cite{openclip}. 

\begin{figure*}[t!]
    \centering
    \includegraphics[width=\linewidth]{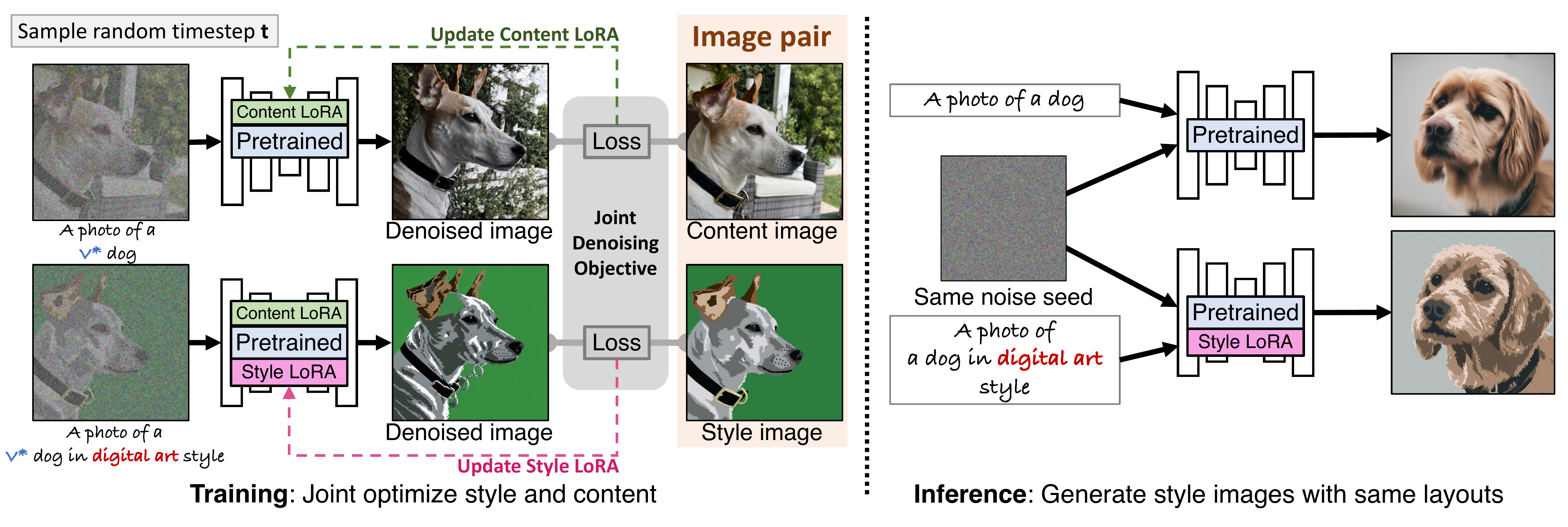}
        \vspace{-5pt}
    \caption{\textbf{Method overview}. (Left) We disentangle style and content from an image pair by jointly training two low-rank adapters, StyleLoRA and ContentLoRA, representing style and content, respectively. Our training objective consists of two losses: The first loss fine-tunes ContentLoRA to reconstruct content image conditioned on a content prompt. The second loss encourages reconstructing the style image using \textit{both} StyleLoRA and ContentLoRA conditioned on a style prompt, but we only optimize Style LoRA for this loss. (Right) At inference time, we only apply StyleLoRA to customize the model. Given the same noise seed, the customized model generates a stylized counterpart of the original pre-trained model output. {\menlo V*} is a fixed random rare token that is a prompt modifier for the content image. Style image credits: \href{https://www.instagram.com/parkhouse_art/}{Jack Parkhouse}}
    \label{fig:method}
        \vspace{-10pt}
\end{figure*}

\myparagraph{Model customization with Low-Rank Adapters.}
Low-Rank Adapters (LoRA)\cite{hu2022lora} is a parameter-efficient fine-tuning method~\cite{houlsby2019parameter} that applies low-rank weight changes $\Delta\theta_{\text{LoRA}}$ to pre-trained model weights $\theta_0$. For each layer with an initial weight $W_0 \in \mathbb{R}^{m \times n}$, the weight update is defined by $\Delta W_{\text{LoRA}} = BA$, a product of learnable matrices $B \in \mathbb{R}^{m \times r}$ and  $A \in \mathbb{R}^{r \times n}$, where $r \ll \min(m, n)$ to enforce the low-rank constraint. The weight matrix of a particular layer with LoRA is:
\begin{equation}
    \begin{aligned}
        W_{\text{LoRA}} = W_0 + \Delta W_{\text{LoRA}} = W_0 + BA.
    \end{aligned}
    \label{eqn:lora}
\end{equation}

At inference time, the LoRA strength is usually controlled by a scaling factor $\alpha \in [0, 1]$ applied to the weight update $\Delta W_{\text{LoRA}}$~\cite{dreamboothlora}:
\begin{equation}
    \label{eq:oldinference}
    \begin{aligned}
        W_{\text{LoRA}} = W_0 + \alpha \Delta W_{\text{LoRA}}.
    \end{aligned}
\end{equation}
LoRA has been applied for customizing text-to-image diffusion models to learn new concepts with as few as three to five images~\cite{dreamboothlora}.

\subsection{Style Extraction from an image pair}\label{sec:analogy_customization}

We aim to customize a pre-trained model with an artistic style in order to stylize the original model outputs while preserving their content, as shown in Figure~\ref{fig:method} (right). To achieve this, we introduce style LoRA weight $\theta_\text{style} = \theta_0 + \Delta\theta_\text{style}$. While a pre-trained model generates content from a noise seed and text $c$, style LoRA's goal is to generate a stylized counterpart of original content from the same noise seed and a style-specific text prompt $\c_\text{style}$, where $\c_\text{style}$ is original text $c$ appended by suffix {\menlo ``in \wordstyle style''}. Here, {\menlo \wordstyle} is a placeholder for some worded description of the style (e.g., ``digital art''), and style LoRA $\theta_\text{style}$ associates {\menlo \wordstyle} to the desired style.

\begin{figure*}
    \centering
    \includegraphics[width=.9\textwidth]{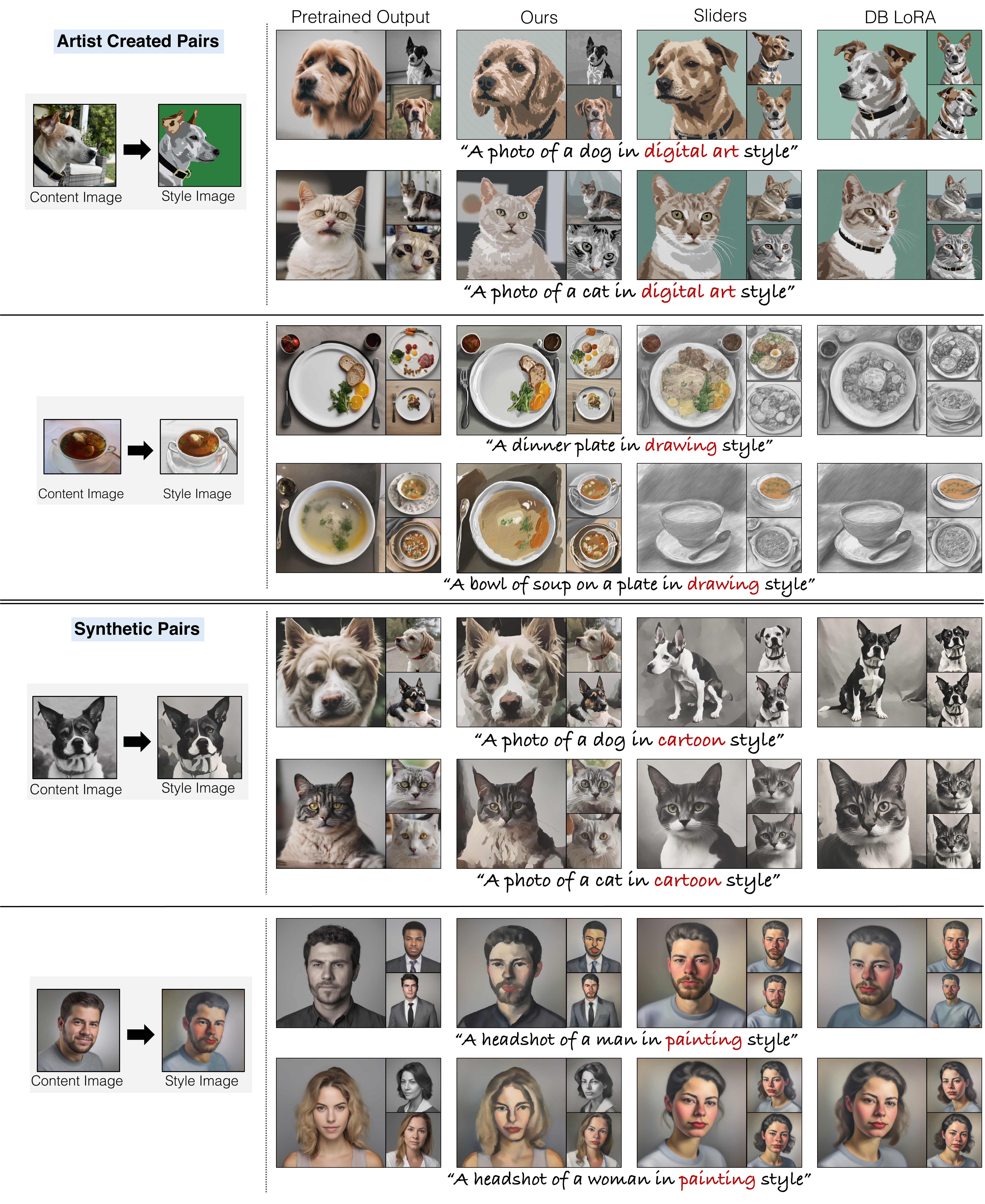}
    \caption{Result of our method compared to the strongest baselines.
    \maxwellchange{When only training with the style image as in DB LoRA, the image structure is not preserved, and overfitting occurs. While Concept Slider's training scheme~\cite{gandikota2023concept} uses both style and content images, it still exhibits overfitting and loss of structure in many cases.} Our method preserves the structure of the input image while faithfully applying the desired style. We use style guidance strength $3$ and classifier guidance strength $5$.
    Style image credits: \href{https://www.instagram.com/parkhouse_art/}{Jack Parkhouse} (First row) and \href{https://www.instagram.com/aaronhertzmann/?hl=en}{Aaron Hertzmann} (Second row)}
    \label{fig:main_result}
\end{figure*}

Unfortunately, learning style LoRA $\theta_\text{style}$ from a single style image often leads to copying content (Figure~\ref{fig:main_result}). Hence, we explicitly learn disentanglement from a style and content image, denoted by $\x_\text{style}$ and $\x_\text{content}$, respectively.

\myparagraph{Disentangling style and content.}
We leverage the fact that the style image shares the same layout and structure as the content image. Our key idea is to learn a separate content LoRA $\theta_\text{content} = \theta_0 + \Delta\theta_\text{content}$ to reconstruct the content image. By explicitly modeling the content, we can train the style LoRA to ``extract'' the stylistic differences between the style and content image. We apply both style and content LoRA to reconstruct the style image, i.e., $\theta_\text{combined} = \theta_0 + \Delta\theta_\text{content} + \Delta\theta_\text{style}$.
This approach prevents leaking the content image to style LoRA, resulting in a better stylization model. %

During training, we feed the content LoRA $\theta_\text{content}$ with a content-specific text $\c_\text{content}$, which contains a random rare token {\menlo V*}, and feed the combined model $\theta_\text{combined}$ with $\c_\text{style}$, where $\c_\text{style}$ is {\menlo ``$\{\c_\text{content}\}$ in \wordstyle style''}. 
Figure~\ref{fig:method} (Left) summarizes our training process.

\myparagraph{Jointly learning style and content.}
We employ two different objectives during every training step. To learn the content of the image, we first employ the standard training objective for diffusion models as described in Section \ref{sec:prelim} with the content image:
\begin{equation}
    \label{eq:content}
    \begin{aligned}
        \mathcal{L}_{\text{content}} = \mathbb{E}_{\epsilon,\x_\text{content},t}\left[w_t\|\epsilon - \epsilon_{\theta_\text{content}}(\x_{t, \text{content}}, \c_\text{content}, t)\|^2\right],
    \end{aligned}
\end{equation}
where $\epsilon_{\theta_\text{content}}$ is the denoiser with content LoRA applied, $\x_{t, \text{content}}$ is a noisy content image at timestep $t$, and $\c_\text{content}$ is text representing the content image, including some rare token {\menlo V*}. Next, we optimize the combined style and content weights to reconstruct the style image. In particular, we only \maxwellchange{train} the style LoRA \maxwellchange{weights} during this step, while stopping the gradient flow to the content LoRA \maxwellchange{weights} via stopgrad $\text{sg}[\cdot]$: %
\begin{equation}
    \label{eq:combination}
    \begin{aligned}
        \theta_{\text{combined}} = \theta_0 + \text{sg}[\Delta \theta_{\text{content}}] + \Delta \theta_{\text{style}}.
    \end{aligned}
\end{equation}

We then apply diffusion objective to train $\theta_\text{combined}$ to denoise $\x_\text{t,style}$, a noisy style image at timestep $t$:
\begin{equation}
    \label{eq:combined}
    \begin{aligned}
        \mathcal{L}_{\text{combined}} = \mathbb{E}_{\epsilon,\x_\text{style},t}\left[w_t\|\epsilon - \epsilon_{\theta_\text{combined}}(x_{t, \text{style}}, \c_\text{style}, t)\|^2\right], 
    \end{aligned}
\end{equation}
where $\epsilon_{\theta_\text{combined}}$ is the denoiser with both LoRAs applied as in Equation \ref{eq:combination}, $\c_\text{style}$ is {\menlo ``$\{\c_\text{content}\}$ in \wordstyle style''},
and {\menlo \wordstyle} is a worded description of the style (e.g., ``digital art''). Finally, we jointly optimize the LoRAs with the two losses:
\begin{equation}
    \label{eq:combined2}
    \begin{aligned}
        \min_{\Delta \theta_{\text{content}}, \Delta \theta_{\text{style}}} \mathcal{L}_{\text{content}} + \mathcal{L}_{\text{combined}}
    \end{aligned}
\end{equation}
Figure \ref{fig:method} provides an overview of our method. Next, we discuss the regularization that promotes the disentanglement of style from content.

\begin{figure}[t!]
    \centering
    \includegraphics[width=\linewidth]{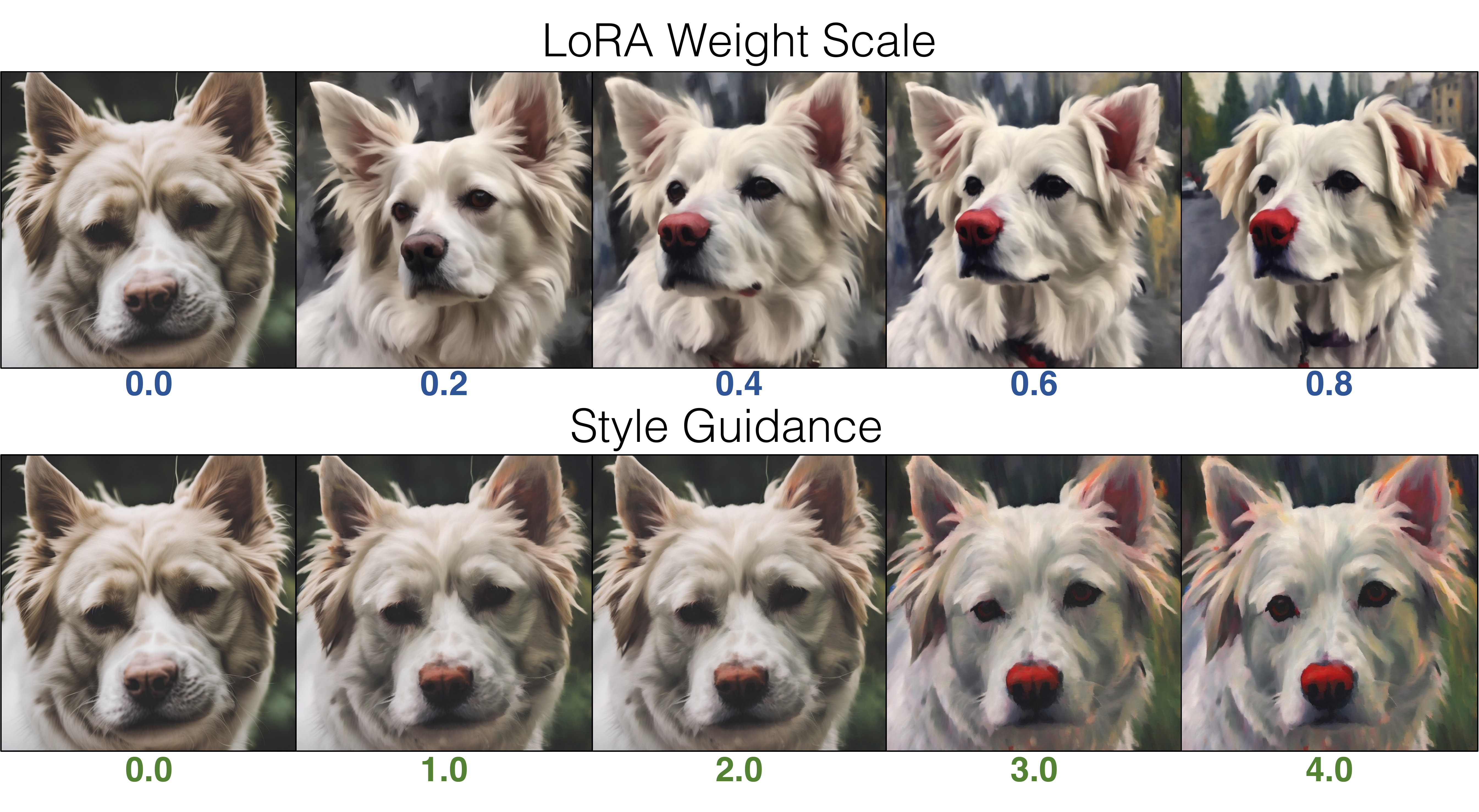}
    \vspace{-20pt}
    \caption{\textbf{Style guidance.} We compare our style guidance and standard LoRA weight scaling~\cite{dreamboothlora}. Style guidance better preserves content when the style is applied. \textbf{{\color{MyDarkBlue} Blue}} and \textbf{{\color{MyDarkGreen} green}} stand for the LoRA weight scale and style guidance scale, respectively. More details of style guidance formulation are in Section~\ref{sec:cfgstyle}.}
    \label{fig:ablation_guidance}

\end{figure}

\begin{figure}[t!]
    \centering
    \includegraphics[width=\linewidth]{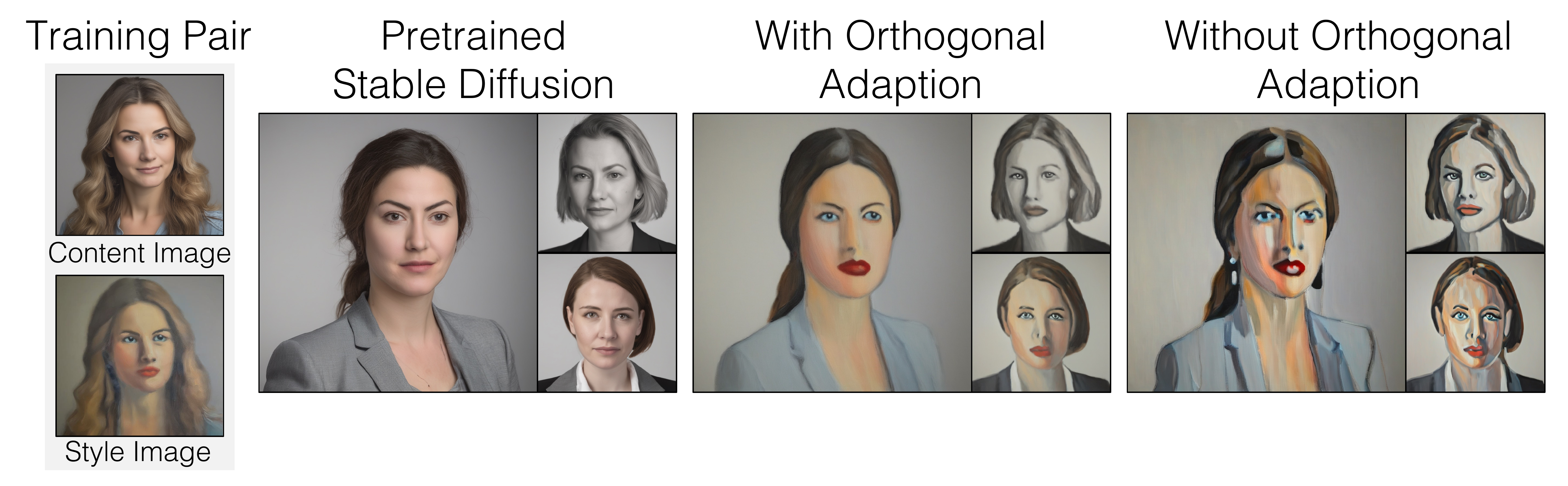}
    \vspace{-20pt}
    \caption{\textbf{Orthogonal adaptation.} Enforcing row-space orthogonality between style and content LoRA improves image quality, where the images capture the style better and have fewer visual artifacts. }
    \label{fig:ablation_orthog}
    \vspace{-8pt}
\end{figure}

\myparagraph{Orthogonality between style and content LoRA.}
\label{sec:orthog}
To further encourage style and content LoRAs to represent separate concepts, we enforce orthogonality upon the LoRA weights. We denote by $W_0$ the original weight matrix and $W_\text{content}$, $W_\text{style}$ the LoRA modifications (layer index omitted for simplicity). Reiterating Equation~\ref{eqn:lora}, we decompose $W_\text{content}$, $W_\text{style}$ into low-rank matrices:
\begin{equation}
    \begin{aligned}
        W_\text{content} =& W_0 + B_\text{content}A_\text{content} \\ \;W_\text{style} =& W_0 + B_\text{style}A_\text{style}.
    \end{aligned}
\end{equation}

We initialize $B_\text{content}, B_\text{style}$ with the zero matrix and choose the rows of $A_\text{content}$, $A_\text{style}$ from an orthonormal basis. We then fix $A_\text{content}$, $A_\text{style}$ and only update $B_\text{content}$, $B_\text{style}$ in training. This forces the style and content LoRA updates to respond to orthogonal inputs, and empirically reduces visual artifacts, as shown in Figure~\ref{fig:ablation_orthog}. This technique is inspired by Po et al.~\cite{po2023orthogonal}. While their work focuses on merging multiple customized objects after each is trained separately, we apply the method for style-content separation during joint training.

\subsection{Style Guidance}
\label{sec:cfgstyle}
A common technique to improve text-to-image model's sample quality is via classifier-free guidance~\cite{ho2022classifier}:
\begin{equation}
    \label{eq:cfg}
    \begin{aligned}
\hat{\epsilon}_\theta(\x_t, \mathbf{c}) = \epsilon_\theta(\x_t, \varnothing) + \lambda_{\text{cfg}} (\epsilon_\theta(\x_t,\mathbf{c}) - \epsilon_\theta(\x_t, \varnothing)),
    \end{aligned}
\end{equation}
where $\hat{\epsilon}_\theta(\x_t, \mathbf{c}, t)$ is the new noise prediction, $\varnothing$ denotes no conditioning, and $\lambda_{\text{cfg}}$ controls the amplification of text guidance.  For notation simplicity, we omit the timestep $t$ in this equation and subsequent ones.  %

To improve pairwise consistency between original and stylized content, we propose an inference algorithm that preserves the original denoising path while adding controllable style guidance:
\begin{equation}
\label{eq:newinference}
    \begin{aligned}
         \hat{\epsilon}_{\theta_0,\theta_\text{style}}&(\x_t, \c, \c_\text{style}) \\
         = \; & \epsilon_{\theta_0}(\x_t, \varnothing) \\
        +\ & \lambda_{\text{cfg}} (\epsilon_{\theta_0}(\x_t, \c) - \epsilon_{\theta_0}(\x_t, \varnothing)) \\
       +\ & \lambda_{\text{style}}(\epsilon_{\theta_{\text{style}}}(\x_t, \c_{ \text{style}}) - \epsilon_{\theta_0}(\x_t, \c)),
    \end{aligned}
\end{equation}
where style guidance is the difference in noise prediction between style LoRA and the pre-trained model. Style guidance strength is controlled by $\lambda_{\text{style}}$, and setting $\lambda_{\text{style}} = 0$ is equivalent to generating original content. In Figure~\ref{fig:ablation_guidance}, we compare our style guidance against scaling LoRA weights (Equation~\ref{eq:oldinference}), and we find that our method better preserves the layout. 
More details and a derivation of our style guidance are in 
Appendix~\ref{sec:styleguid}.

\secondmaxwellchange{
Previous works have also used multiple guidance terms with diffusion models, including guidance from multiple text prompts using the same model~\cite{liu2022compositional} and additional image conditions~\cite{brooks2023instructpix2pix}. 
Unlike these, we obtain additional guidance from a customized model and apply it to the original model. StyleDrop~\cite{sohn2023styledrop} considers a similar formulation with two guidance terms but for masked generative transformers. SINE~\cite{zhang2023sine} uses a customized content model to apply text-based image editing to a single image, like adding snow. However, we use a customized style model to generate any image with the desired style.}

\myparagraph{Blending multiple learned styles.}
With a collection of models customized by our method, we can blend the learned styles as follows. Given some set of styles $\mathcal{S}$ and strengths $\lambda_{\text{style}_0}, \dots, \lambda_{\text{style}_n}$, we blend the style guidance from each model, and our new inference path is represented by 
\begin{equation}
    \begin{aligned}
         \hat{\epsilon}_{\theta_0,\theta_\text{style}}&(\x_t, \c, \c_\text{style}) \\
         = \; & \epsilon_{\theta_0}(\x_t, \varnothing) \\
        +\ & \lambda_{\text{cfg}} (\epsilon_{\theta_0}(\x_t, \c) - \epsilon_{\theta_0}(\x_t, \varnothing)) \\
        +\ & \sum_{\text{style}_i \in \mathcal{S}}\lambda_{\text{style}_i}(\epsilon_{\theta_{\text{style}_i}}(\x_t, \c_{ \text{style}_i}) - \epsilon_{\theta_0}(\x_t, \c)),
    \end{aligned}
    \label{eqn:blend}
\end{equation}
We can vary the strengths of any parameter $\lambda_{\text{style}_i}$ to seamlessly increase or decrease style application while preserving content. \reffig{blending} gives a qualitative example of blending two different styles while preserving image content. 

\myparagraph{Implementation details.} We train all models using an AdamW optimizer \cite{loshchilov2018decoupled} and learning rate $1\times 10^{-5}$. For baselines, we train for $500$ steps. For our method, we first train our content weights on the content image for $250$ steps, and then train jointly for $500$ additional steps. All image generation is performed using $50$ steps of a PNDMScheduler~\cite{liu2022pseudo}. For all methods using inference with LoRA adapters, we use SDEdit~\cite{meng2021sdedit} to further preserve structure. Specifically, normal classifier-free guidance on the original prompt without style is used for the first 10 steps. We then apply style guidance/LoRA scale for the rest of the timesteps. %

\begin{figure*}[!t]
    \centering
    \includegraphics[width=.87\textwidth]{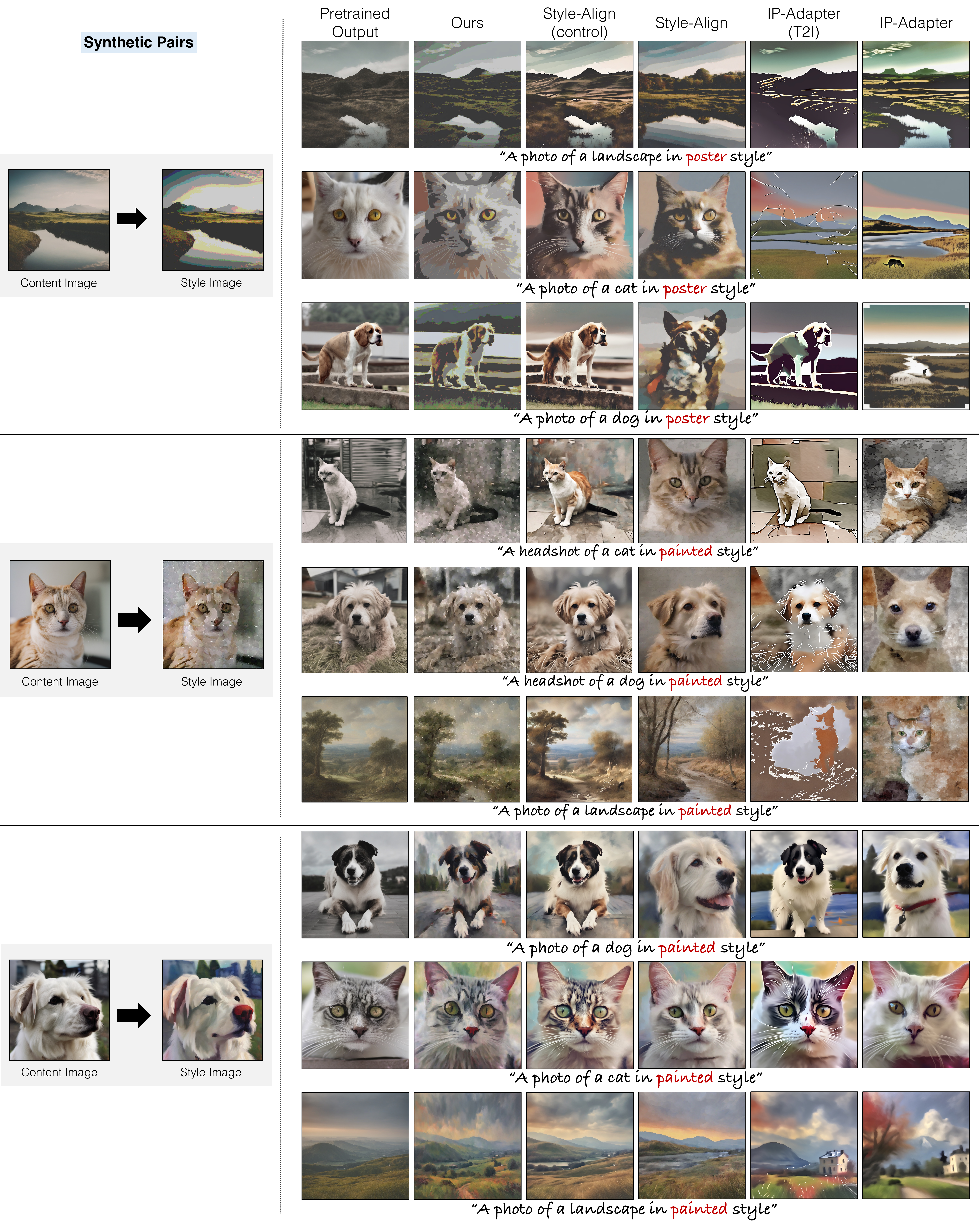}
    \vspace{-8pt}
    \caption{Result of our method compared to the methods without finetuning (zoom in for best viewing). For the baseline methods, we also add the edge map from the pretrained output as an extra condition ($3^{\text{rd}}$ and $5^{\text{th}}$ column). Without this edge map, other methods tend to lose the structure of the pretrained output. In some cases, however, an additional edge map can overly constrain the output of a model, like in the second image pair example. Our method preserves the structure of the Stable Diffusion image while faithfully applying the desired style. We use style guidance strength $3$ and classifier guidance strength $5$ for our method and set the IP-adapter scale and style-alignment scale to $0.5$. }
    \label{fig:main_result_supp}
\end{figure*}

\section{Experiments}
\label{sec:exp}
\subsection{Dataset}
\label{sec:dataset}
In this section, we show our method's results on various image pairs and compare them with several baselines. We explain our dataset, baselines, and metrics in detail, then we present quantitative and qualitative results.

\begin{figure}[t!]
    \centering
    \includegraphics[width=\linewidth]{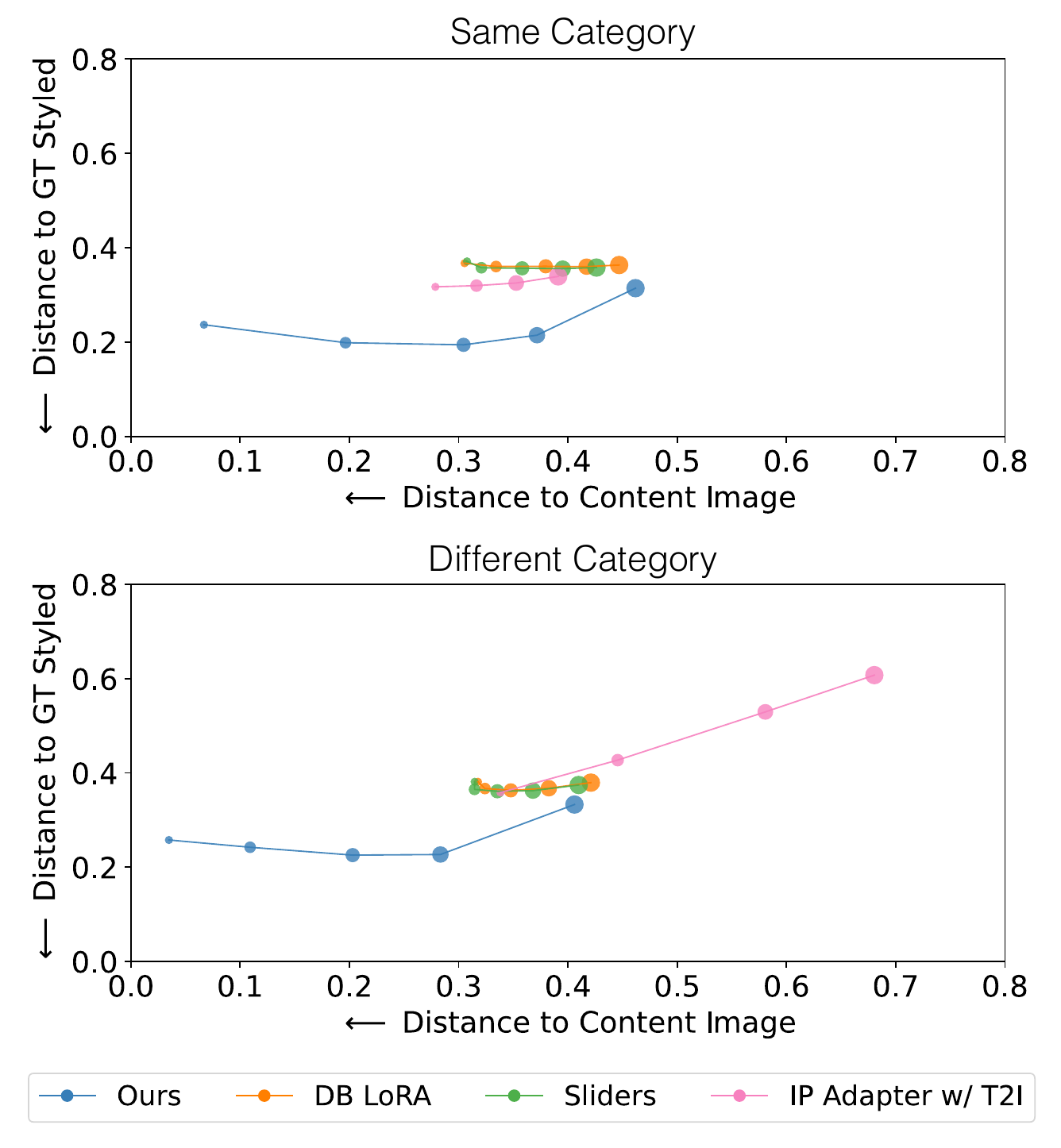}
    \vspace{-20pt}
    \caption{\textbf{Quantitative comparison with baselines on learned style.} Given a fixed inference path, our method pareto dominates baselines for image generation both on the same category as training (left) and when evaluated on categories different from training, e.g., trained on human portraits but tested on dog images (right). 
    We further evaluate the diversity of generated images in the supplement. We show that baselines often lose diversity, while our method leads to diverse generations while still achieving lower perceptual distance to the ground truth style.
    Increased marker size corresponds to an increase in the guidance scale.
    }
    \label{fig:tradeoff_plot}
\vspace{-15pt}
\end{figure}

\myparagraph{Datasets.}
To enable large-scale quantitative evaluation, we construct a diverse set of paired style and content images as follows. %
First, we generate 40 content images for each class: headshots, animals, and landscapes. When generating images in the headshot class, we generate 20 images with the prompt ``{\menlo A professional headshot of a man}'' and 20 images with the prompt \\ ``{\menlo A professional headshot of a woman}''. Similarly, we split the animal class into photos of dogs and cats. To curate synthetic pairs, we then apply image editing or image-to-image translation methods to all the content images to obtain the stylized version. For each unique prompt, we choose a \textit{single paired instance} as training data and hold out the other pairs with the same prompt as a test set (Same Category). For each prompt, we also choose 5 pairs from each of the other prompts as a secondary test set (Different Category). We show all our synthetic training image pairs in 
\refapp{implementation_details}.
By leveraging synthetic pairs for evaluation, we can train on a single synthetic pair and test our results against held out synthetic style images. Secondly, we qualitatively compare against single artist pairs in Figure \ref{fig:main_result}.
Next, we describe the specific methods to create the paired dataset. First, we consider the diffusion based image editing technique LEDITS++\cite{brack2023ledits++} to translate images into paintings. Next, we consider Cartoonization \cite{wang2020learning}, a GAN-based translation technique that aplies a cartoon-like effect. We also consider Stylized Neural Painting \cite{zou2021stylized}, which turns photos into painting strokes using a rendering based approach. Finally, we consider the image filtering technique posterization. We provide a more detailed description of each method for creating synthetic pairs in \refapp{implementation_details}.

\subsection{Baselines and Evaluation Metrics}
\label{sec:baselines}
\myparagraph{Baselines.}
We compare our method against -- (1) DreamBooth LoRA~\cite{hu2022lora,dreamboothlora} (DB LoRA), (2) Concept Sliders~\cite{gandikota2023concept} (3) IP-adapters~\cite{ye2023ip}, (4) IP-adapters w/ T2I, and (5) StyleDrop \cite{sohn2023styledrop}. DB LoRA uses only the style image and fine-tunes low-rank adapters in all the linear layers in the attention blocks of the diffusion model. We evaluate different amounts of style applications for DB LoRA using the standard LoRA scale~\cite{dreamboothlora}
. Concept sliders presents a paired image model customization method that trains a single low-rank adapter jointly on both images, with different reconstruction losses for the style and content images. We also evaluate using the standard LoRA scale
. IP-adapters is an encoder-based method that does not require training for every style and takes a style image as an extra condition separate from the text prompt. Increasing or decreasing the guidance from the input style image is possible by scaling the weight of the image conditioning. We consider the SDXL~\cite{podell2023sdxl} implementation of this method. For the IP-Adapter, we compare against the stronger baseline of providing extra conditioning of an edge map of the content image through T2I Adapters \cite{mou2024t2i} to preserve the content image structure. The recently proposed Styledrop \cite{sohn2023styledrop} technique for learning new styles is based on MUSE \cite{chang2023muse}, and uses human feedback in its method. Since MUSE is not publicly available, we follow Style-Aligned Image Generation's \cite{hertz2023style} setup, and implement a version of StyleDrop on SDXL. Specifically, we train low-rank linear layers following each Feed-Forward layer in the attention blocks of SDXL. For a fair comparison, we train Styledrop without human feedback.

\myparagraph{Evaluation metrics.}
When evaluating the performance of each method, we consider two quantitative metrics: perceptual distance to ground truth style images and structure preservation from the original image. A better customization method will have a low perceptual distance to the ground truth style images while still preserving content of the original image before adding style.
 We measure these using -- (1) \textit{Distance to GT Styled}: given holdout ground truth style images, we measure the perceptual distance between our styled outputs and the ground truth style images using DreamSim~\cite{fu2023dreamsim}, a recent method for measuring the perceptual distance between images. DreamSim image embeddings are comprised of an ensemble of image embedding models, including CLIP~\cite{radford2021learning} and DINO~\cite{caron2021emerging}, which are then fine-tuned so the final embeddings respect human perception.
We measure DreamSim distance as (1 - cosine similarity between DreamSim embeddings), where a lower value implies that the images are perceptually more similar. (2) \textit{Distance to Content Image}: to measure content preservation after style application, we measure the perceptual distance of our generated style image to the original content image with no style guidance. We again use DreamSim, this time comparing styled and content images. Note here that a perceptual distance of zero to the content image is undesirable, as this would require no style to be applied. However, a better-performing method should obtain a better tradeoff between the two distances. (3) We also perform a \textit{human preference study} of our method against baselines.

\begin{figure}[t!]
\centering
\includegraphics[width=\linewidth]{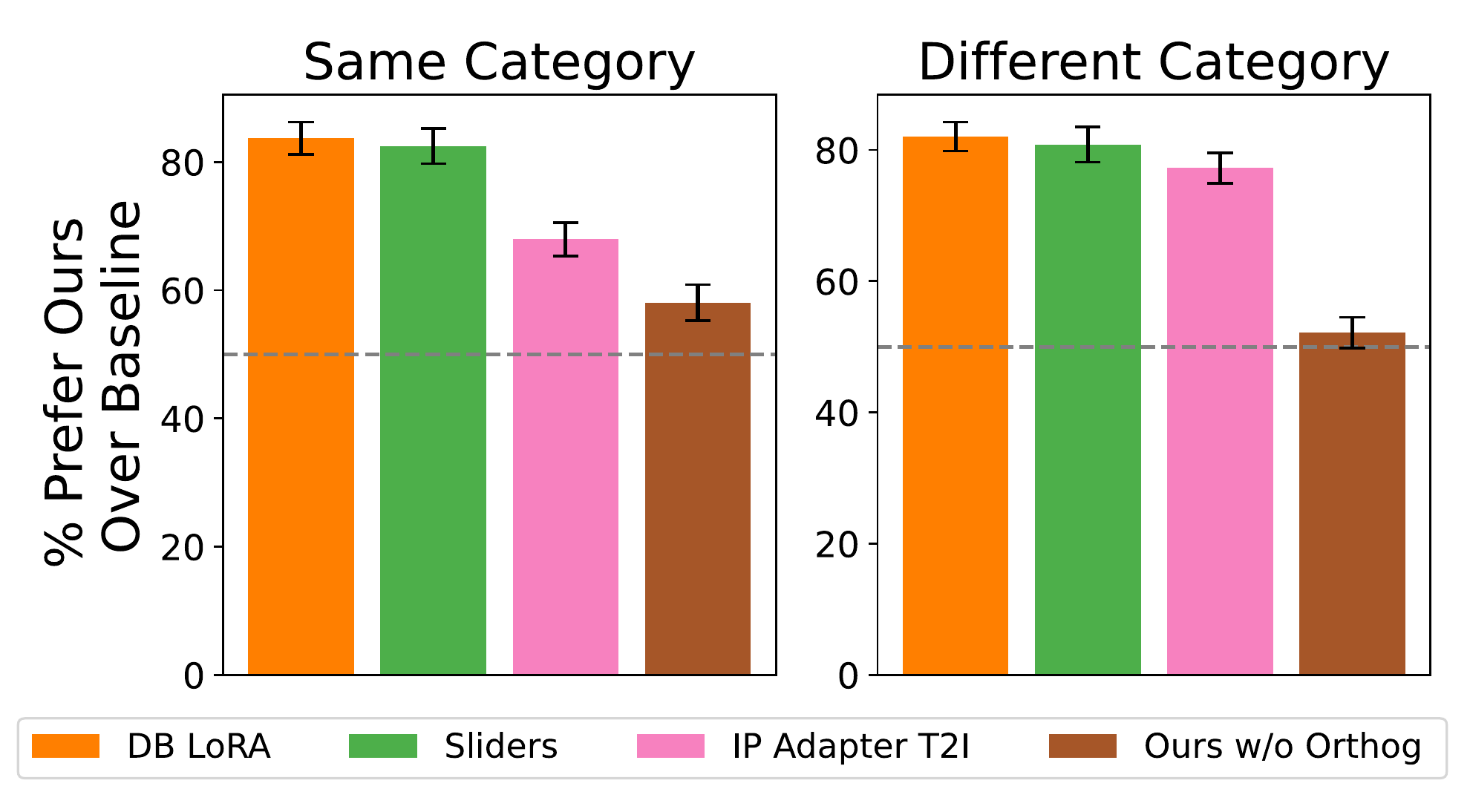}
\vspace{-15pt}
\caption{\textbf{Human preference study.} Our method is preferred over the baselines ($\geq 60\%$). Further, our full method, including orthogonal weight matrices (Section \ref{sec:orthog}), is preferred over the one w/o orthogonal weight matrices, specifically for the same category as training pair, e.g., trained on a headshot of a man and tested on other headshots of man.  The Gray dashed line denotes 50\% chance performance.}
\vspace{-15pt}
\label{fig:human_eval}
\end{figure}

\subsection{Results}

\myparagraph{Quantitative evaluation.}
We show quantitative results against the highest performing baselines in Figure \ref{fig:tradeoff_plot}. Increased marker size (circles) indicates the higher application of style, and line color determines the method. When evaluating style similarity vs. structure preservation in Figure \ref{fig:tradeoff_plot}, we see that our training method's Pareto dominates all baselines, yielding lower perceptual distance to style images while still being perceptually similar to the original content image. We find that DB LoRA and Styledrop perform similarly, and report Styledrop results in 
\refapp{ablation}.
Finally, we consider our method with LoRA scale during inference and other baselines with our style guidance during inference for ablation, and provide results in 
\refapp{ablation}.

\myparagraph{Qualitative evaluation.} We compare our method with the highest performing baselines in \reffig{main_result}. \maxwellchange{The finetuning-based methods DB LoRA~\cite{hu2022lora,dreamboothlora} and Concept Sliders ~\cite{gandikota2023concept} outperform the encoder-based method~\cite{ye2023ip} for our task. Hence, we compare against that in \reffig{main_result}. For both baselines, we modulate style application with LoRA scale (Equation \ref{eq:oldinference}).} We observe that DB LoRA often fails to generate the style-transformed version of the original image and overfits to the training pair image when generating similar concepts. There are two main reasons why this may occur. First, we are in a challenging case where there is only $1$ training image instead of the usual $3-5$ images that customization methods use. Second, we are prompting the model on the same or very similar text prompts to the training prompt, and the baseline method overfits to the training image for these prompts. Our method preserves the structure of the original image while applying the learned style. Moreover, applying our style guidance instead of the LoRA scale benefits the baseline method as well (\reffig{main_result}, last $2$ columns), as it can better preserve the structure of the original image, though it still tends to overfit to the content of the training image. We observe a similar issue for other baselines as well.

\maxwellchange{We compare our method to non finetuning-based methods in Figure \ref{fig:main_result_supp}. We observe that these methods perform worse than finetuning-based methods, especially when generating images in a different category to the training style image. We also compare with baselines using our style guidance for style application at inference time in 
\refapp{extra_quant}
.}

\myparagraph{User preference study.} We perform a user preference study using Amazon Mechanical Turk. We test our method against all baselines, as well as a version of our method trained without the orthogonality constraint. Specifically, we test on all datasets in Section \ref{sec:dataset}. When evaluating against DB LoRA and Concept Sliders, we consider inference with both LoRA scale as in Equation \ref{eq:oldinference} and style guidance as in Equation $\ref{eq:newinference}$. For each method, we pick a single style strength that performs most optimally according to quantitative metrics as in Figure \ref{fig:tradeoff_plot}. Full details are available in 
\refapp{implementation_details}, and a user study with all baselines is available in \refapp{extra_quant}
. We collect $400$ responses per paired test of ours vs the other method. The user is shown an image generated via our method and an image generated via the other method and asked to select the image that best applies the given style to the new content image. We provide a detailed setup of the user study 
as well as a user study on baseline methods using our style guidance in \refapp{implementation_details}
As shown in Figure~\ref{fig:human_eval}, our method is favored by users in comparison to baselines, whether evaluating images generated within the same category as the training image pair or across different categories. Secondly, users prefer our full method to ours without the orthogonality constraint, specifically when evaluating on the same category as training.

\begin{figure}
    \centering
    \includegraphics[width=\linewidth]{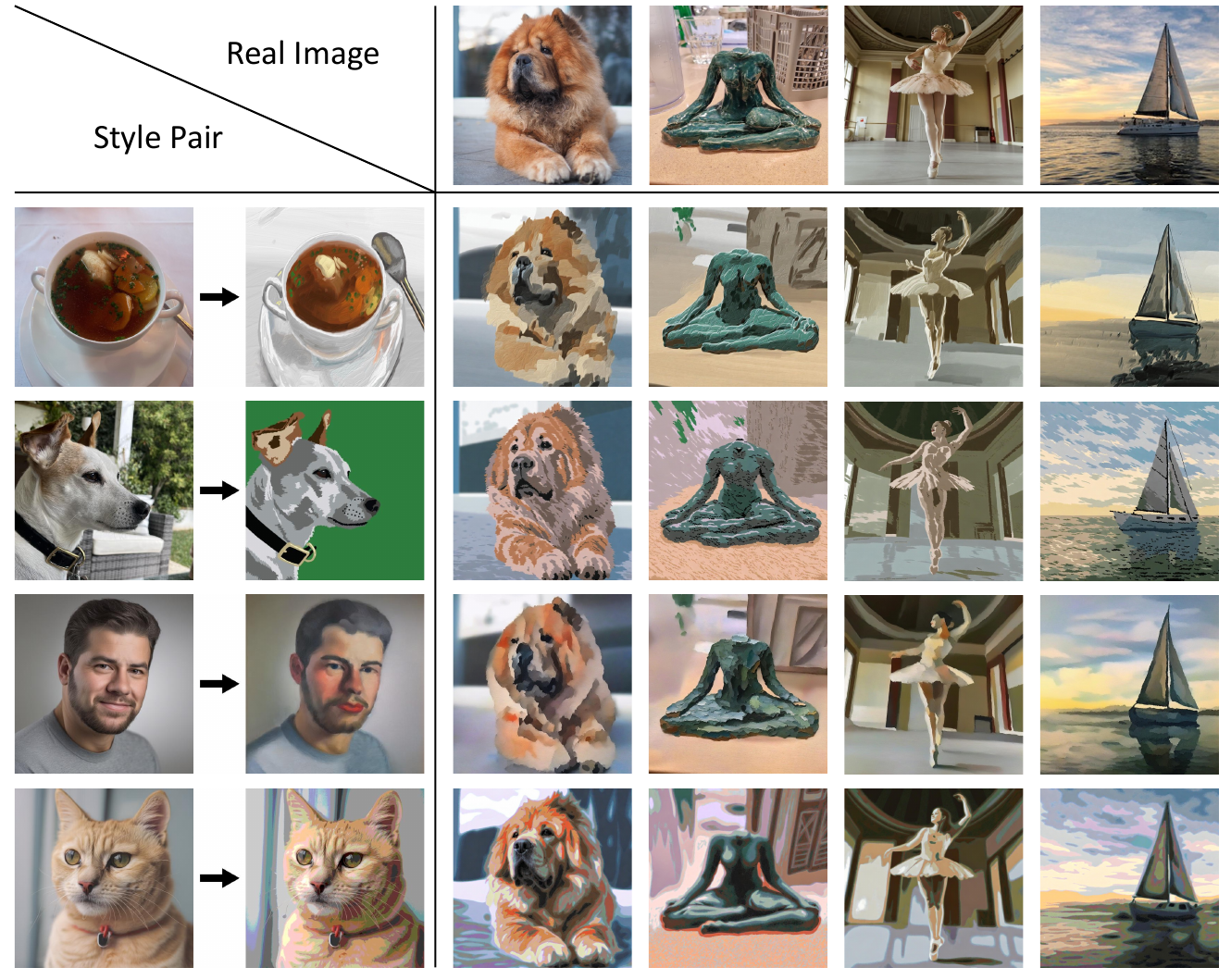}
        \vspace{-15pt}
    \caption{\secondmaxwellchange{\textbf{Real image editing.} We can edit real images by inverting a real image into a noisy latent code using a diffusion inversion pipeline \cite{song2020denoising, garibi2024renoise} and an image prompt $ \c$. From here, we apply style guidance (Equation \ref{eq:newinference}) with the same prompt $\c$ and new style prompt $\c_{\text{style}}$ for the desired style application.}}
    \label{fig:real_editing}
    \vspace{-2pt}
\end{figure}

\myparagraph{Real Image Editing.}
\secondmaxwellchange{
Our method can also stylize real images. 
We use DDIM inversion \cite{song2020denoising,garibi2024renoise} to invert images into their noisy latent codes at some intermediate step using a reference prompt $\c$. From here, we use our style guidance (Equation \ref{eq:newinference}) with reference prompt $\c$ and new prompt $\c_{\text{style}}$ to denoise the noisy latent code to a stylized image. In \reffig{real_editing}, we show real image editing results. We provide more details in Appendix \ref{sec:editing_details}.}

\begin{figure}
    \centering
    \includegraphics[width=\linewidth]{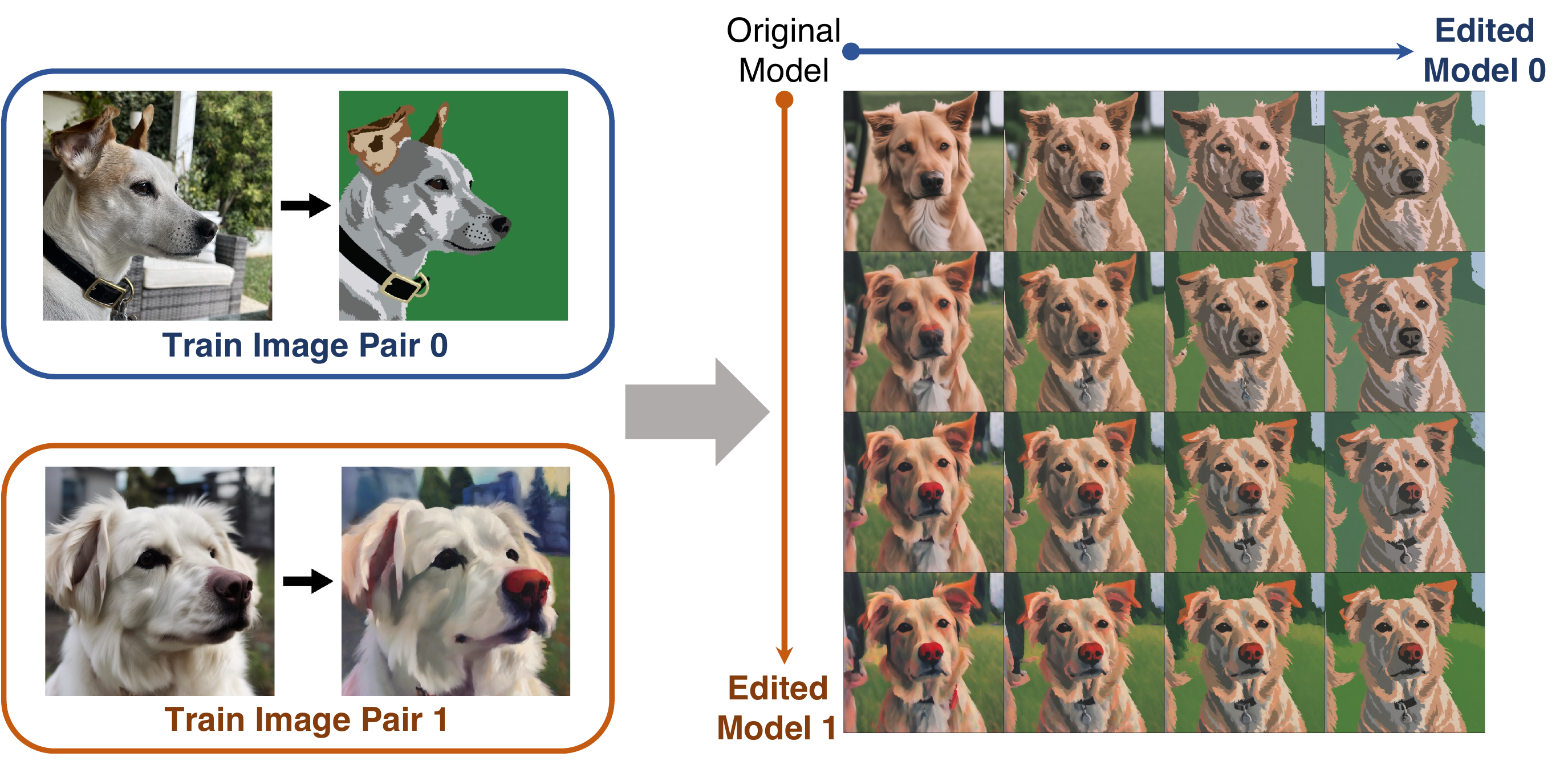}
        \vspace{-15pt}
    \caption{\textbf{Blending multiple style guidances.} We can compose multiple customized models by directly blending each style guidance together. Adjusting the blending strength of each model allows us to acquire a smooth style transition. Each stylized image corresponds to different style guidance strengths. Train Image Pair 0 style image credits: \href{https://www.instagram.com/parkhouse_art/}{Jack Parkhouse}}
    \label{fig:blending}
    \vspace{-2pt}
\end{figure}

\myparagraph{Blending learned styles.}
We show that we can blend the learned styles by applying a new inference path, defined in Equation~\ref{eqn:blend}. In Figure~\ref{fig:blending}, we show the results of blending two models. We can seamlessly blend the two styles at varying strengths while still preserving the content.

\begin{figure}[!t]
    \centering
    \includegraphics[width=\linewidth]{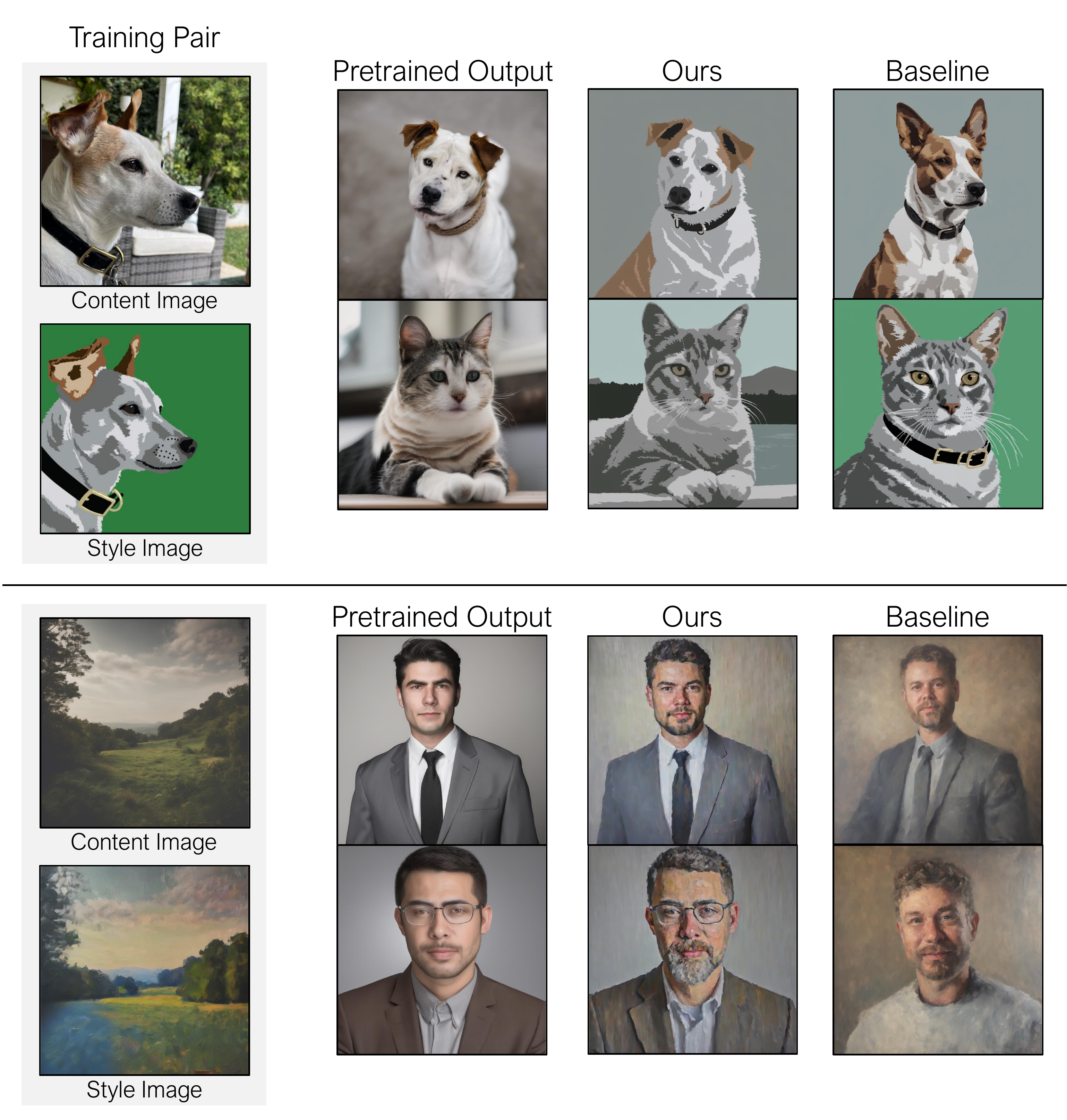}
    \caption{\textbf{Limitations.} \emph{Top}: our method can cause structure changes in some instances, like change of body position or background changes. \emph{Bottom}: 
     Our method can change the content in some cases from pretrained output, like the addition of facial hair. We display Baseline DB LoRA for comparison.
    }
    \label{fig:limitation}
    \vspace{-15pt}
\end{figure}

\section{Discussion and Limitations}
In this work, we have introduced a new task: customizing a text-to-image model with a single image pair. To address this task, we have developed a customization method that explicitly disentangles style and content through both training objectives and a separated parameter space. Our method enables us to grasp the style concept without memorizing the content of input examples. While our approach outperforms existing customization methods, it still exhibits several limitations, as discussed below. 

\myparagraph{Limitations.}
First, our method may occasionally fail to completely maintain input structure, as demonstrated in Figure~\ref{fig:limitation}. This could occur as background/pose change (Top), or as additional features being added (Bottom).

Second, our current method relies on test-time optimization, which takes around 15 minutes on a single A5000 GPU. This can be computationally demanding if we need to process many image styles. Leveraging encoder-based approaches~\cite{arar2023domain,ruiz2023hyperdreambooth} for predicting style and content weights in a feed-forward manner could potentially speed up the customization process. 

\myparagraph{Acknowledgments.}
We would like to thank Ali Jahanian, Gaurav Parmar, Ruihan Gao, Sean Liu, and Or Patashnik for their insightful feedback and input that contributed to the finished work. 
We also thank Jack Parkhouse and Aaron Hertzmann for providing style images.
Maxwell Jones is supported by the Rales Fellowship and the Siebel Scholar fellowship. This project is partly supported by the Amazon Faculty Research Award, NSF IIS-2239076, Open Philanthropy, and the Packard Fellowship.

{\small
\bibliographystyle{ieee_fullname}
\bibliography{main}
}
\clearpage

\appendix

\renewcommand{\thefootnote}{\arabic{footnote}}

\noindent{\Large\bf Appendix}
\vspace{5pt}

In Section \ref{sec:diversity}, we evaluate our method against baselines on the diversity metric, showing that our method leads to more diverse generations comparatively. We also show more qualitative results along with a comparison to the concurrent work of Style Aligned Image Generation~\cite{hertz2023style}. In Section \ref{sec:styleguid}, we then present details of our style guidance formulation. \secondmaxwellchange{In Section \ref{sec:editing_details}, we provide implementation details for real image editing, as well as quantitative and qualitative comparisons of real image editing with our method vs baselines}. Finally, in Section \ref{sec:implementation_details}, we provide more implementation details, including the setup for our human preference study and the full synthetic training dataset used for evaluation.

\section{More Quantitative and Qualitative Results}
\label{sec:ablation}
\begin{figure*}[t!]
    \centering
    \includegraphics[width=\linewidth]{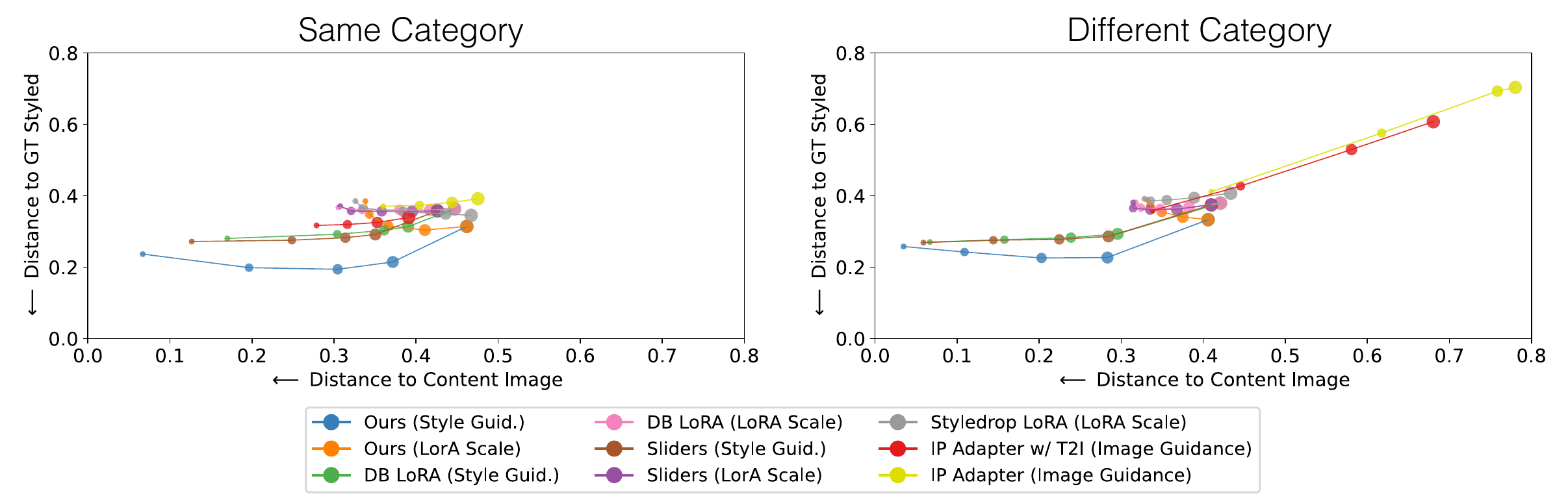}
    \caption{\textbf{Quantitative comparison with baselines on learned style.} Given a fixed inference path, our method's pareto dominates baselines for image generation both on the same category as training (left) and when evaluated on categories different from training, e.g., trained on human portraits but tested on dog images (right). 
    \maxwellchange{Secondly, our proposed style guidance outperforms standard LoRA weight scale guidance for our training method (blue vs. orange), DB LoRA (green vs. pink), and Sliders (brown vs. purple)}. 
    Increased marker size corresponds to an increase in guidance scale.
    }
    \label{fig:tradeoff_plot_large}
\end{figure*}

\begin{figure}[t!]
\centering
\includegraphics[width=\linewidth]{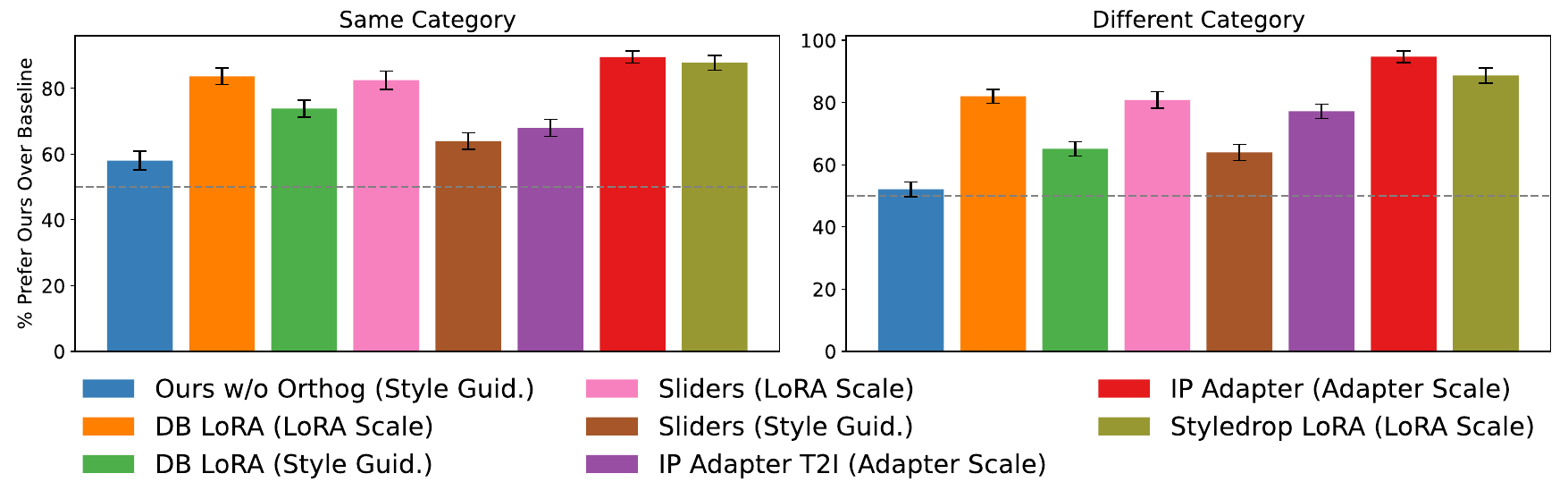}
\caption{\textbf{Human preference study.} Our method is preferred over the baselines ($\geq 60\%$), including those using our style guidance. The Gray dashed line denotes 50\% chance performance.}
\vspace{-15pt}
\label{fig:human_eval_large}
\end{figure}

\myparagraph{Style Guidance Ablation}
We compare our method, DB LoRA \cite{hu2022lora, loraimplementation}, and sliders \cite{gandikota2023concept} with both style guidance and LoRA scale used at inference. We also compare against IP Adapters \cite{ye2023ip} without T2I adapter conditioning, and our Styledrop \cite{sohn2023styledrop} implementation. In Figure \ref{fig:tradeoff_plot_large}, style guidance outperforms the LoRA scale for Ours (blue vs orange), DB LoRA (green vs. pink), and Concept Sliders (brown vs. purple) while our final method still pareto dominates all others, highlighting the effectiveness of both our training scheme and our inference scheme. In Figure \ref{fig:human_eval_large}, our method is still preferred greater than 60 percent of the time against all baselines, while baselines that use style guidance for inference have a higher success rate than those using DB LoRA.

\begin{figure*}[t!]
    \centering
    \includegraphics[width=0.95\linewidth]{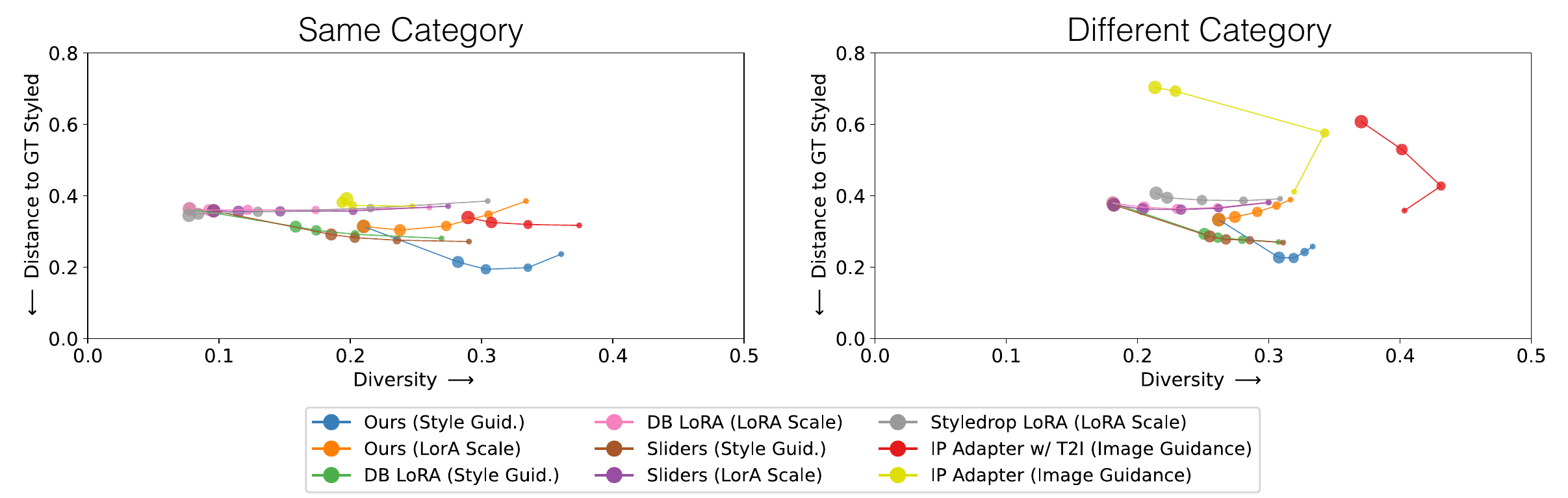}
    \caption{\textbf{Quantitative comparison on Diversity metric.} Our method with style guidance has high diversity and low perceptual distance to ground truth style images both on the same category as training (left) and when evaluated on categories different from training, e.g., trained on human portraits but tested on dog images (right). 
    Methods without edge control tend to lose diversity indicating overfitting, and methods with edge control have similar/higher diversity, but much worse style application.  
    Increased marker size corresponds to an increase in style guidance scale.
    }
    \label{fig:tradeoff_plot_diversity}
\end{figure*}

\label{sec:diversity}
\myparagraph{Diversity metric.}
To measure the overfitting behavior of our method and baselines, we consider a diversity metric. Concretely, we measure the DreamSim \cite{fu2023dreamsim} perceptual distance between any two images trained with the same style image pair and generated with the same prompt. We then average results over training pairs and prompts. More formally, we let
\begin{equation}
    \label{eq:overfitting}
    \begin{aligned}
        & \text{DreamSim } \text{Diversity} \\
        =& \mathbb{E}_{S \in \mathcal{S}, P \in \mathcal{P}}\left[\mathbb{E}_{i_1,i_2 \in \text{data}_{S, P}}\text{DreamSim}(i_1,i_2)\right]
    \end{aligned}
\end{equation}
where $\mathcal{S}$ is the set of style image pairs, $\mathcal{P}$ is the set of prompts, and $\text{data}_{S, P}$ is the set of images generated with prompt $P$ by a model customized on style $S$. $\text{DreamSim}(\cdot, \cdot)$ is DreamSim perceptual distance. A decrease in DreamSim Diversity indicates that all images in a certain domain are becoming perceptually similar, which may indicate overfitting to the style training image. Methods that do not overfit the style training image should have higher diversity scores while also having a low perceptual distance to the ground truth testing style images. We present our findings in Figure \ref{fig:tradeoff_plot_diversity}. Our method is able to achieve a low perceptual distance to style ground-truth images while maintaining higher diversity scores. As shown in Figure \ref{fig:main_result} in the main paper, the baseline results mode collapses to the training image, thus lowering their diversity score as they all become perceptually similar to each other.

\begin{figure*}[!t]
    \centering
    \includegraphics[width=.9\textwidth]{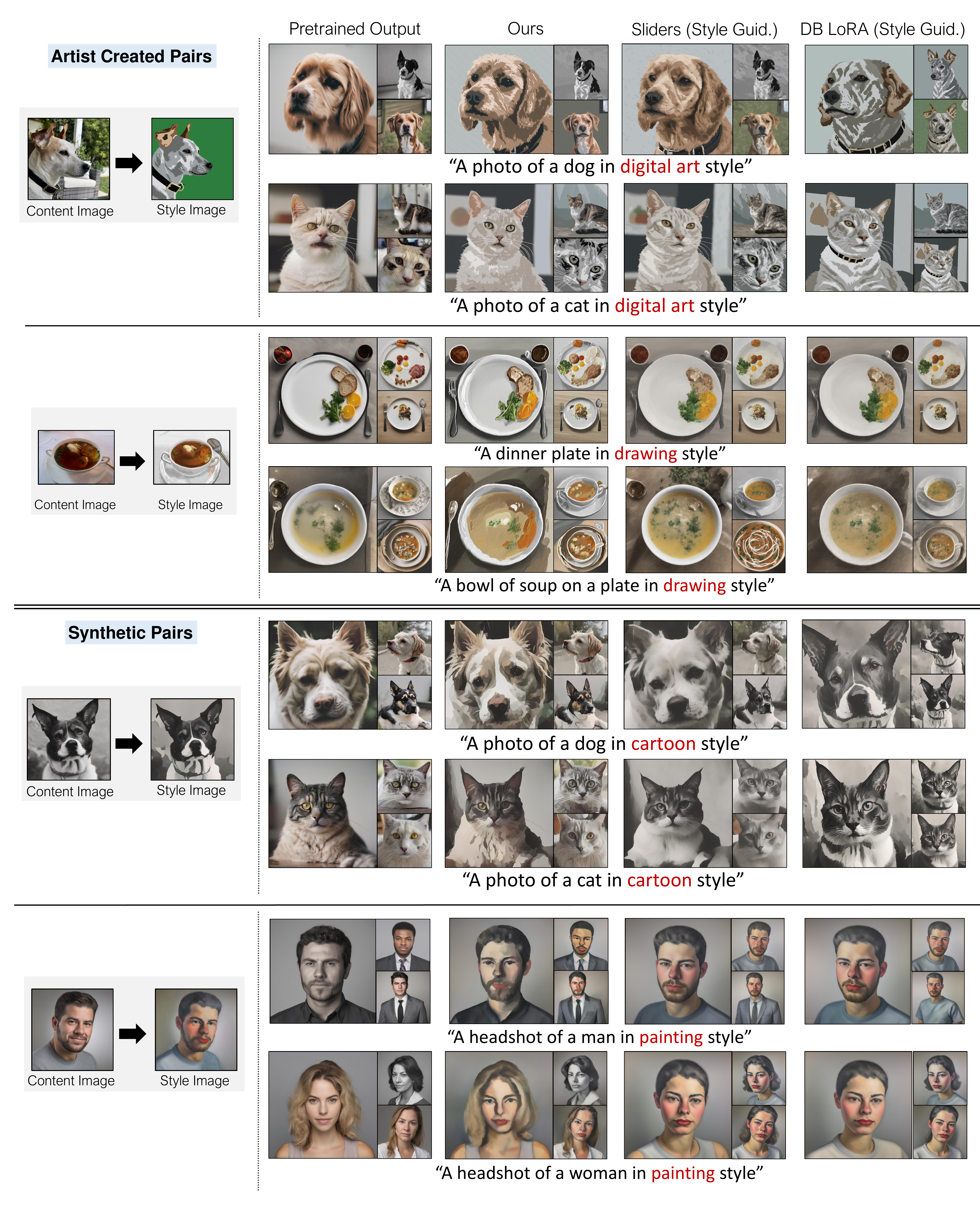}
    \caption{\maxwellchange{Result of our method compared to the strongest baselines, but replacing LoRA scale (Eq. 
    \ref{eq:oldinference}
    ) with our style guidance (Eq. 
    \ref{eq:newinference}
    ) for the baselines. While our style guidance increases baseline performance over LoRA scale images displayed in Figure 
    \ref{fig:main_result}
    , our method is still superior in terms of preserving content while applying style.}}
    \label{fig:baseline_style_guid}
        \vspace{-10pt}
\end{figure*}

\begin{figure*}[t!]
    \centering
    \includegraphics[width=0.95\linewidth]{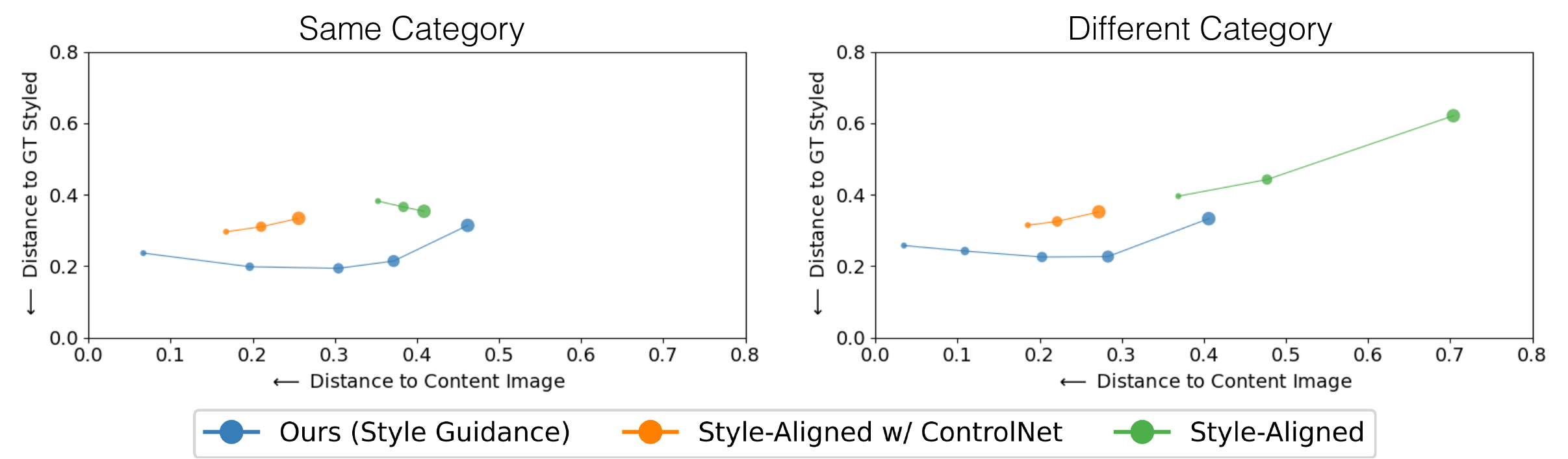}
    \caption{\textbf{Style similarity with Style Aligned \cite{hertz2023style}.} Our method Pareto dominates both versions of Style Aligned Image Generation for image generation both on the same category as training (left) and when evaluated on categories different from training, e.g., trained on human portraits but tested on dog images (right).
    }
    \label{fig:tradeoff_plot_style_aligned}
\end{figure*}

\begin{figure*}[t!]
    \centering
    \includegraphics[width=0.95\linewidth]{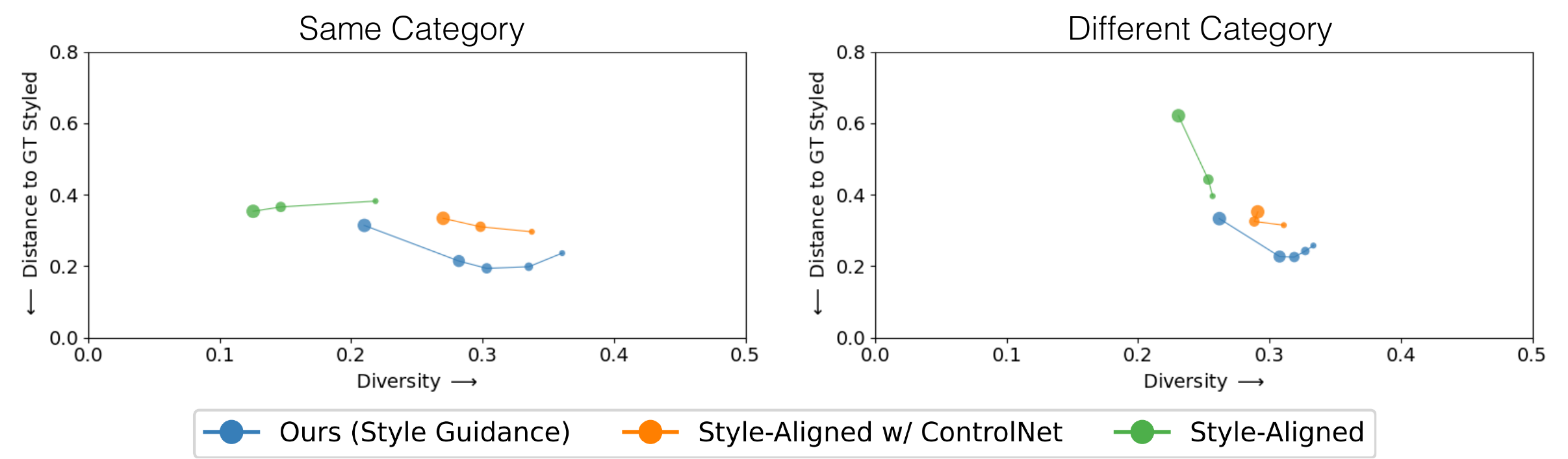}
    \caption{\textbf{Image diversity with Style Aligned \cite{hertz2023style} on learned style (Diversity).} Our method has high diversity and low perceptual distance to ground truth style images both on the same category as training (left) and when evaluated on categories different from training, e.g., trained on human portraits but tested on dog images (right) as compared to both versions of Style Aligned Image Generation. 
    }
    \label{fig:tradeoff_plot_diversity_style_aligned}
    \vspace{10pt}
\end{figure*}

\myparagraph{Style Aligned Image Generation \cite{hertz2023style} Baseline}
\label{sec:stylealigned}
Style Aligned Image Generation \cite{hertz2023style} is a recent work for zero-shot style-consistent image generation from an exemplar style image. Given the exemplar style image, it is first inverted to a noise map; then for a new text prompt, the image is generated by attending to both its own self-attention map and the self-attention map from the style exemplar at every denoising step. 
We compare against this baseline by using the style image in our training image pair as an exemplar and generating a new style image with a new text prompt using this method.
Optionally, we condition this generation on the edge map of the newly generated image without attention sharing using ControlNet~\cite{zhang2023adding} to help with content preservation. We show the qualitative results of our method compared to all the variants of this baseline in Figure 
\ref{fig:main_result_supp}
 of the main paper. Figures \ref{fig:tradeoff_plot_style_aligned} and \ref{fig:tradeoff_plot_diversity_style_aligned} show quantitative comparison, where our method outperforms this baseline in terms of both style similarity and diversity metric. We achieve lower perceptual distance to the style ground-truth images, low perceptual distance from content images, and high diversity.

\myparagraph{Extra Qualitative Evaluation}
\label{sec:extra_quant}
\maxwellchange{We compare our method with the highest-performing baselines, but use our style guidance (Equation 
10 of the main paper
) to apply stylization during inference for these baselines. We present our results in Figure \ref{fig:baseline_style_guid}. First, we notice that using style guidance for adding style allows the baseline methods to better preserve original content over LoRA scale (Figure \ref{fig:baseline_style_guid} vs Figure 
\ref{fig:main_result}
 in the main paper
). While adding our style guidance is better able to preserve content while applying style for baseline methods, our full method is still able to outperform baselines with style guidance applied.}

\section{Style Guidance Details}
\label{sec:styleguid}
In this section, we derive our style guidance formulation. We consider the probability of latent $\x$ with multiple conditionings~\cite{brooks2023instructpix2pix}, i.e., the text prompt $c_t$ and a class of style images $c_\text{style}$. First, we apply Bayes' rule:
\begin{equation}
    \begin{aligned}
        & P(\x|c_t, c_\text{style}) \\
        =&  \frac{P(\x, c_t, c_\text{style})}{P(c_t, c_\text{style})} \\
        =& \frac{P(c_\text{style}|c_t, \x)P(c_t|\x)P(\x)}{P(c_t, c_\text{style})}
    \end{aligned}
\end{equation}
Applying logarithm on both sides, we get: 
\begin{equation}
    \begin{aligned}
        & \log(P(\x|c_t, c_\text{style})) \\
        &= \log(P(c_\text{style}|c_t, \x)) + \log(P(c_t|\x))
        + \log(P(\x)) \\ & \ \ \ \ - \log(P(c_t, c_\text{style}))
    \end{aligned}
\end{equation}
Next, we take the derivative with respect to $\x$:

\begin{equation}
    \begin{aligned}
        & \nabla_\x\log(P(\x|c_t, c_\text{style}))  \\
         = & \nabla_\x\log(P(c_\text{style}|c_t, \x)) 
        + \nabla_\x\log(P(c_t|\x)) 
        + \nabla_\x\log(P(\x)) \\
    =& \ \nabla_\x\log\left(\frac{P(c_\text{style},c_t, \x)}{P(c_t, \x)}\right) \\
    &+ \nabla_\x\log\left(\frac{P(c_t,\x)}{P(\x)}\right) \\
    &+ \nabla_\x\log\left(P(\x)\right) \\
    = & \ (\nabla_\x\log P(c_\text{style},c_t, \x) - \nabla_\x\log P(c_t, \x)) \\
        &+ (\nabla_\x\log P(c_t,\x) - \nabla_\x\log P(\x)) \\
        &+ (\nabla_\x\log\left(P(\x)\right))
    \end{aligned}
    \label{eq:logmath}
\end{equation}
As usual, we approximate $\nabla_\x\left(\log P(c_t, \x)\right)$ via $\epsilon_\theta(\x_t, c_t)$ and $\nabla_\x\log\left(P(\x)\right)$ via $\epsilon_\theta(\x_t, \varnothing)$. \textbf{Importantly, we approximate}
\begin{equation}
    \label{eq:updaterulemath}
    \begin{aligned}
\nabla_\x \log(P(\x_t|c_t, \c_\text{style})) \approx \epsilon_{\theta_{\text{style}}}(\x_t, c_{t, \text{style}})
    \end{aligned}
\end{equation}
 where $c_t$ is the original text prompt, $\c_\text{style}$ is the class of stylized images from the training style, $\theta_{\text{style}}$ is the UNet with style LoRA adapters applied, and $c_{t, \text{style}} = $ {\menlo ``$\{\c_t\}$ in \wordstyle style''}. Here, we use $c_t$ to push the prediction in the text direction, and both text conditioning ({\menlo ``in \wordstyle style''}) and low-rank adapters ($\theta_{\text{style}}$) to push the prediction into the class of images in the artist's style denoted by $c_{\text{style}}$. Following this, our new score estimate is:
\begin{align}
    \notag \hat{\epsilon}_\theta&(\x_t, c_t, \c_\text{style}) \\
    = & \epsilon_\theta(\x_t, \varnothing) \\
    &+  \lambda_{\text{cfg}} (\epsilon_\theta(\x_t, c_t) - \epsilon_\theta(\x_t, \varnothing)) \\
  \notag  &+  \lambda_{\text{style}}(\epsilon_{\theta_{\text{style}}}(\x_t, c_{t, \text{style}}) - \epsilon_{\theta}(\x_t, c_t)) \\
\end{align}
$\lambda_{\text{cfg}}$ and $\lambda_{\text{style}}$ are guidance scales that can be varied as in classifier-free guidance~\cite{ho2022classifier}. %
Given a fixed $\lambda_{\text{cfg}}$, we can vary the $\lambda_{\text{style}}$ term as desired to generate an original guidance $\lambda_{\text{cfg}}$ image with varying amounts of style. Notice that at $\lambda_\text{cfg} = \lambda_{\text{style}}$, the $ \epsilon_\theta(\x_t, \varnothing)$ terms cancel and we are left with the original classifier-freeguidance. 

\section{Real Image Editing Details}\label{sec:editing_details}

\begin{figure}[t!]
    \centering
    \includegraphics[width=\linewidth]{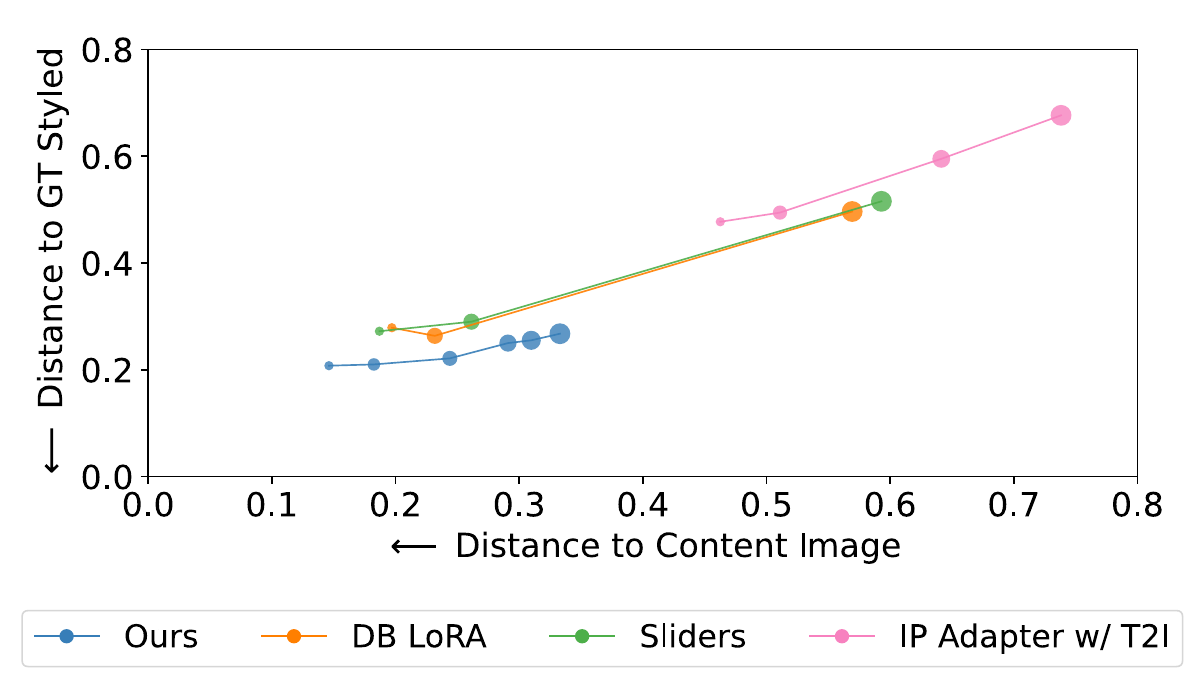}
   \caption{\secondmaxwellchange{\textbf{Quantitative comparison with baselines on real image editing.} Given a fixed classifier-free guidance scale, our method pareto dominates baselines for image generation. During inference, we use style guidance for our method, classifier free guidance for DB LoRA and Sliders, and Image Guidance for IP Adapter w/ T2I. 
    Increased marker size corresponds to an increase in guidance scale.
    }}
    \label{fig:tradeoff_real_editing}
\end{figure} 

\begin{figure*}
    \centering
    \includegraphics[width=.8\linewidth]{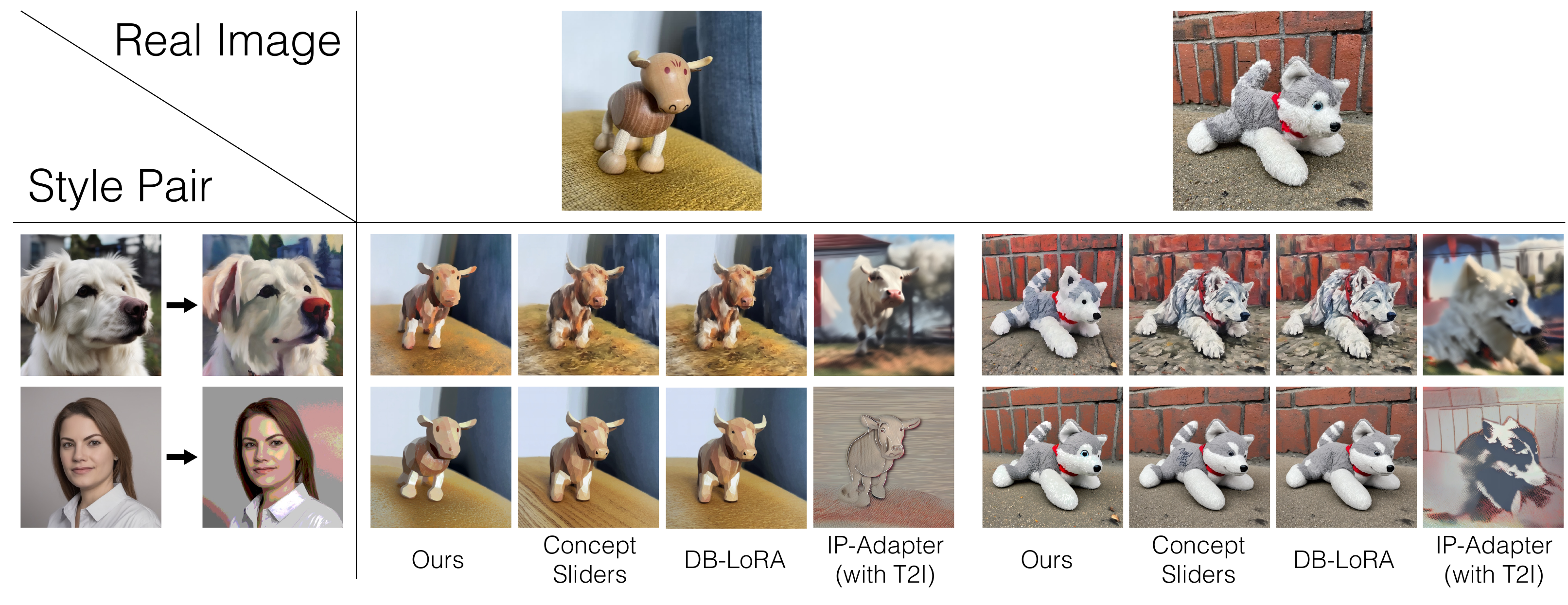}
        \vspace{-15pt}
    \caption{\secondmaxwellchange{\textbf{Real image editing comparison.} We compare our real image editing results to other baselines. Our method is able to best preserve content from the real image while applying style, across both training style pairs and input images. We set classifier-free guidance to 5 in all cases. For our method, we set our style guidance (Eq. 10 in the main body) to 6. For baselines, we set LoRA scale (Eq. 3 in the main body) to 1. We invert all images using DDIM inversion \cite{song2020denoising, garibi2024renoise}.}}
    \label{fig:real_editing_comparison}
    \vspace{-2pt}
\end{figure*}

\secondmaxwellchange{
We provide a quantitative and qualitative analysis of our real image editing results compared to baselines. We provide quantitative results in Figure~\ref{fig:tradeoff_real_editing}. For quantitative evaluation, we use a set of 17 real images from the B-LoRA dataset \cite{frenkel2024implicit} and add style with all trained style models (see Figure \ref{fig:training_data} for all training pairs). We measure the perceptual distance to content and style images with DreamSim~\cite{fu2023dreamsim}. Our method Pareto dominates other baselines, yielding both lower perceptual distance to style images and lower perceptual distance to the original content image. We provide qualitative results against baselines in Figure \ref{fig:real_editing_comparison}. Our method is able to better preserve the real image structure while applying the style from the training pair. We invert the noise to 600 steps for our method and 700 steps for baselines, as we find experimentally that baselines do not apply style when real images are only inverted to 600 steps. We use ReNoise Inversion \cite{garibi2024renoise} for our DDIM inversion implementation on SDXL.}

\section{Implementation Details}\label{sec:implementation_details}

\begin{figure*}[!t]
    \centering
    \includegraphics[width=\textwidth]{figure_pdf/training_data.pdf}
    \caption{\textbf{Training Data.} We present the synthetic training data set used for evaluation, where each pair is used as a single training instance. Each column corresponds to a different style, and each row corresponds to a different content category.}
    \label{fig:training_data}
\end{figure*}

We describe the specific methods used to create the paired dataset in detail.

\myparagraph{LEDITS++}\cite{brack2023ledits++} is a diffusion-based image editing technique that transforms an image by updating the inference path of a diffusion model. After fine-grained inversion, a global prompt and a set of translation prompts representing a new style or object are used to perform the image translation. We leverage LEDITS++ on all images with the translation prompt ``Impressionist style''. Further, we change the word ``photo'' to ``painting'' in the original prompt when generating the style image.  

\myparagraph{White-box cartoonization.}
Cartoonization \cite{wang2020learning} is a GAN-based image-to-image translation technique that applies a cartoon-like effect to real images. We apply the cartoonized model to our set of generated images to create image pairs. 

\myparagraph{Stylized neural painting.}
Stylized Neural Painting \cite{zou2021stylized} is a rendering based image to image translation technique where an image is reconstructed via $N$ painting strokes, where the strokes are guided by a loss function that encourages the final translated image to resemble the original. We use the Neural Painting model with $N = 1000$ to create image pairs.

\myparagraph{Posterization.}
Posterization is an image filtering technique that reduces the number of distinct colors in a given image to some fixed number $N$, reducing color variation and creating fixed color areas. We apply posterization to images in our training set with $N = 8$.

\myparagraph{Training data.}
We present our full training set of $20$ different style transformations in Figure~\ref{fig:training_data}. Each image pair is a standalone training instance used in our method. We consider four different styles (posterization, impressionist, neural painting, cartoonization), with each column corresponding to a single style. For each style, we consider five categories for training (man, woman, dog, cat, landscape).  

\secondmaxwellchange{
\myparagraph{Style descriptor Ablation}
We consider replacing a text style descriptor (i.e., ''digital art style'') with a random rare token (i.e., ''$S*$ style''). Applying the same style strength, we quantitatively find that using words yields better distance to content and GT style image ($0.340\pm0.03$, $0.194\pm0.03$) than using rare tokens ($0.348\pm0.04$, $0.209\pm0.03$), averaged over all test cases. These experiments confirm the remark in StyleDrop~\cite{sohn2023styledrop} that using a text style descriptor produces better results. 
}

\myparagraph{Mechanical Turk details.}
\begin{figure*}[!t]
    \centering
    \includegraphics[width=.7\textwidth]{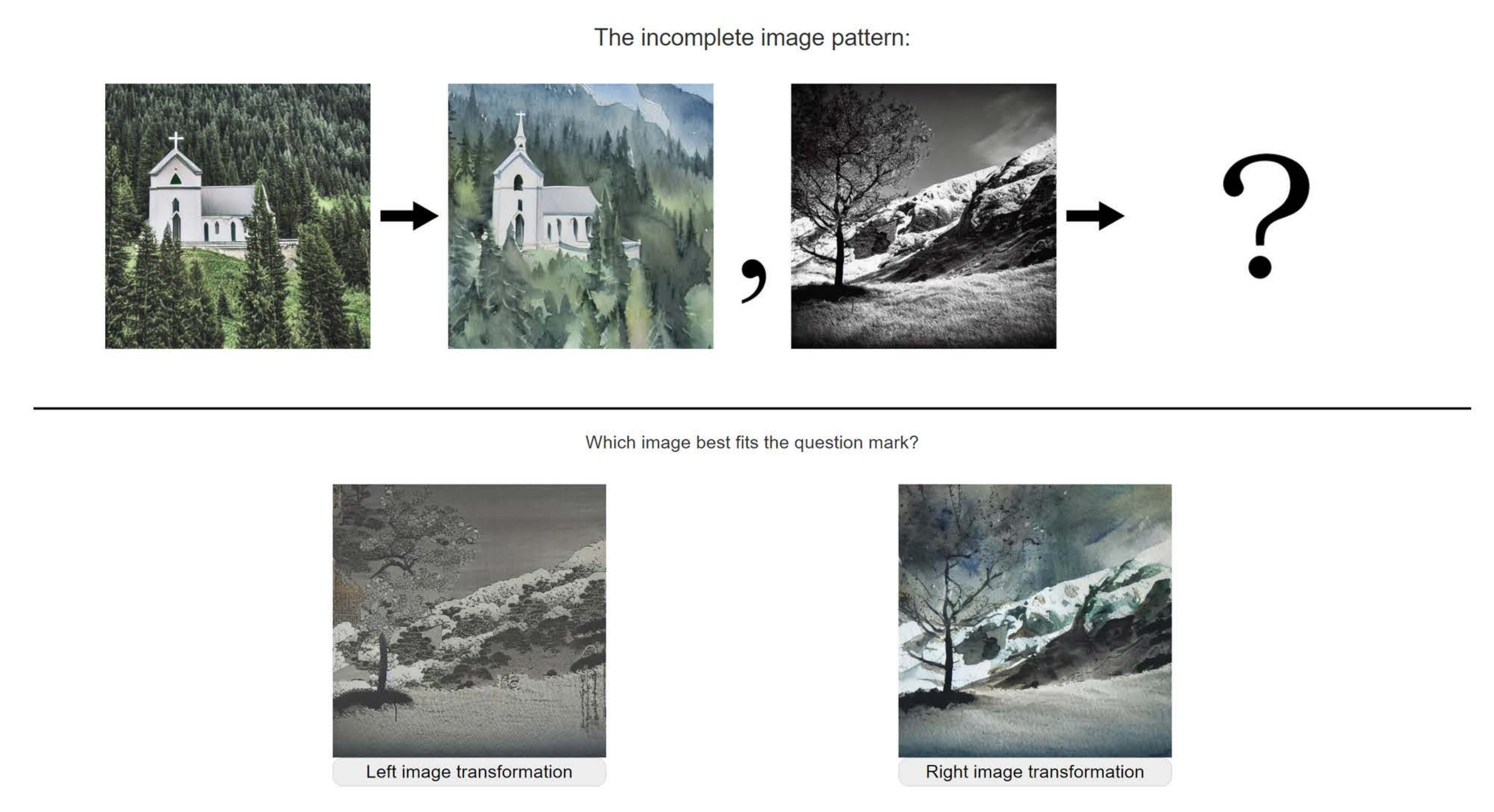}
    \caption{Mturk User Interface
    }

    \label{fig:mturk}
        \vspace{-10pt}
\end{figure*}

When running Amazon Mechanical Turk, we prompt users with an analogy-style interface. First, we provide the training pair of images, followed by the testing content image, and two options for possible styled examples. After viewing both images, users choose either the left or right image. Figure \ref{fig:mturk} shows an example. Each individual user is presented with four training examples, as in Figure \ref{fig:mturk}, followed by 16 random testing examples comparing our method with one of our baselines. We survey 75 users for each of the 16 individual studies and use bootstrapping to obtain variance estimates. In total, we collect 19200 user samples. \maxwellchange{For each method, we pick a stylization hyperparameter based on 
\reffig{tradeoff_plot_large}
. For details, see Table \ref{tab:hyperparams}}

\begin{table}[!t]
\centering
\setlength{\tabcolsep}{5pt}
\resizebox{\linewidth}{!}{
\begin{tabular}{l c c}
\toprule
\multicolumn{1}{c}{Method} & \multicolumn{2}{c}{Hyperparameter value} \\ \cmidrule{2-3} 
\multicolumn{1}{c}{}                        & Same Category    & Different Category    \\
\midrule
Ours (Style Guid.)    & 3 & 4 \\
Ours w/ Orthog (Style Guid.)    & 3 & 4 \\
DB LoRA (Style Guid.)   & 2 & 4   \\
DB LoRA (LoRA Scale)   & 0.4 & 0.8 \\
Concept Sliders (Style Guid.)  & 2 & 4  \\
Concept Sliders (LoRA Scale)  & 0.6 & 0.8\\
StyleDrop LoRA (LoRA Scale)   & 0.6 & 1 \\
IP Adapter w/T2I (Image Guidance)   & 0.5 & 0.5   \\
IP Adapter (Image Guidance)   & 0.5 & 0.5 \\
\bottomrule
\end{tabular}
}
\caption{\textbf{Experiment Hyperparameters.} \maxwellchange{We choose a fixed stylization hyperparameter for our own model and each baseline when generating images for Mechanical Turk. When picking a hyperparameter, we try and optimize tradeoffs between style application and content preservation, informed by \reffig{tradeoff_plot_large}. Our style guidance (Equation 
10 in the main body
) generally takes values from 0 to $\lambda_{\text{cfg}} = 5$, while all other stylization hyperparameters generally take values 0 to 1. }
}
\label{tab:hyperparams}
\vspace{-18pt}
\end{table}

\end{document}